\pdfoutput=1
\documentclass[10pt,twocolumn,letterpaper]{article}
\usepackage{wacv}
\usepackage{times}
\usepackage{epsfig}
\usepackage{graphicx}
\usepackage{amsmath}
\usepackage{amssymb}
\usepackage{booktabs}
\usepackage{multirow}
\usepackage{makecell}
\def\wacvPaperID{853} % Enter the WACV Paper ID here

\wacvalgorithmstrack   % Uncomment this line if you are submitting to the Algorithms Track.
\wacvfinalcopy % *** Uncomment this line for the final submission

\usepackage[breaklinks=true, bookmarks=false]{hyperref}
\hypersetup{
  colorlinks   = true, %Colours links instead of ugly boxes
  urlcolor     = blue, %Colour for external hyperlinks
  linkcolor    = red, %Colour of internal links
  citecolor   = green %Colour of citations
}

\usepackage{subcaption}
\begin{document}

%%%%%%%%% TITLE
\title{Patch-level Gaze Distribution Prediction for Gaze Following}

\author{Qiaomu Miao \qquad Minh Hoai \qquad Dimitris Samaras\\
Stony Brook University, Stony Brook, NY 11794, USA\\
{\tt\small \{qiamiao, minhhoai, samaras\}@cs.stonybrook.edu} 
%\and
}
\maketitle
\thispagestyle{empty}

%%%%%%%%% ABSTRACT
\begin{abstract}
   Gaze following aims to predict where a person is looking in a scene, by predicting the target location, or indicating that the target is located outside the image. Recent works detect the gaze target by training a heatmap regression task with a pixel-wise mean-square error (MSE) loss, while formulating the in/out prediction task as a binary classification task. This training formulation puts a strict, pixel-level constraint in higher resolution on the single annotation available in training, and does not consider annotation variance and the correlation between the two subtasks. To address these issues, we introduce the patch distribution prediction (PDP) method. We replace the in/out prediction branch in previous models with the PDP branch, by predicting a patch-level gaze distribution that also considers the outside cases. Experiments show that our model regularizes the MSE loss by predicting better heatmap distributions on images with larger annotation variances, meanwhile bridging the gap between the target prediction and in/out prediction subtasks, showing a significant improvement in performance on both subtasks on public gaze following datasets. Code is available at: \url{https://github.com/qiaomu-miao/gazefollowing_pdp}
\end{abstract}

%%%%%%%%% BODY TEXT
\section{Introduction} \label{intro}

Gaze behavior is an important human behavior that serves a crucial role in inferring social intent and interactions \cite{emery2000eyes,abele1986functions,pfeiffer2012eyes}, assisting human-computer interaction \cite{morimoto2005eye,majaranta2014eye}, predicting learning outcomes \cite{Rebello-etal-PERC19}, and diagnosis of psychological disorders like autism \cite{charman1997infants,jaswal2019being,baron1997mindblindness}. Therefore, analyzing human gaze automatically has attracted significant interest from computer vision researchers. Specifically, gaze following seeks to understand human gaze behavior in the wild by predicting the gaze target of a person inside a scene image in a third-person view, by locating the gaze target if it is located within the image, or indicating that the target is located outside. 

\begin{figure}[t]
\centering
\begin{subfigure}{0.47\linewidth}
    \centering
    \includegraphics[width=\linewidth, height=1.15in]{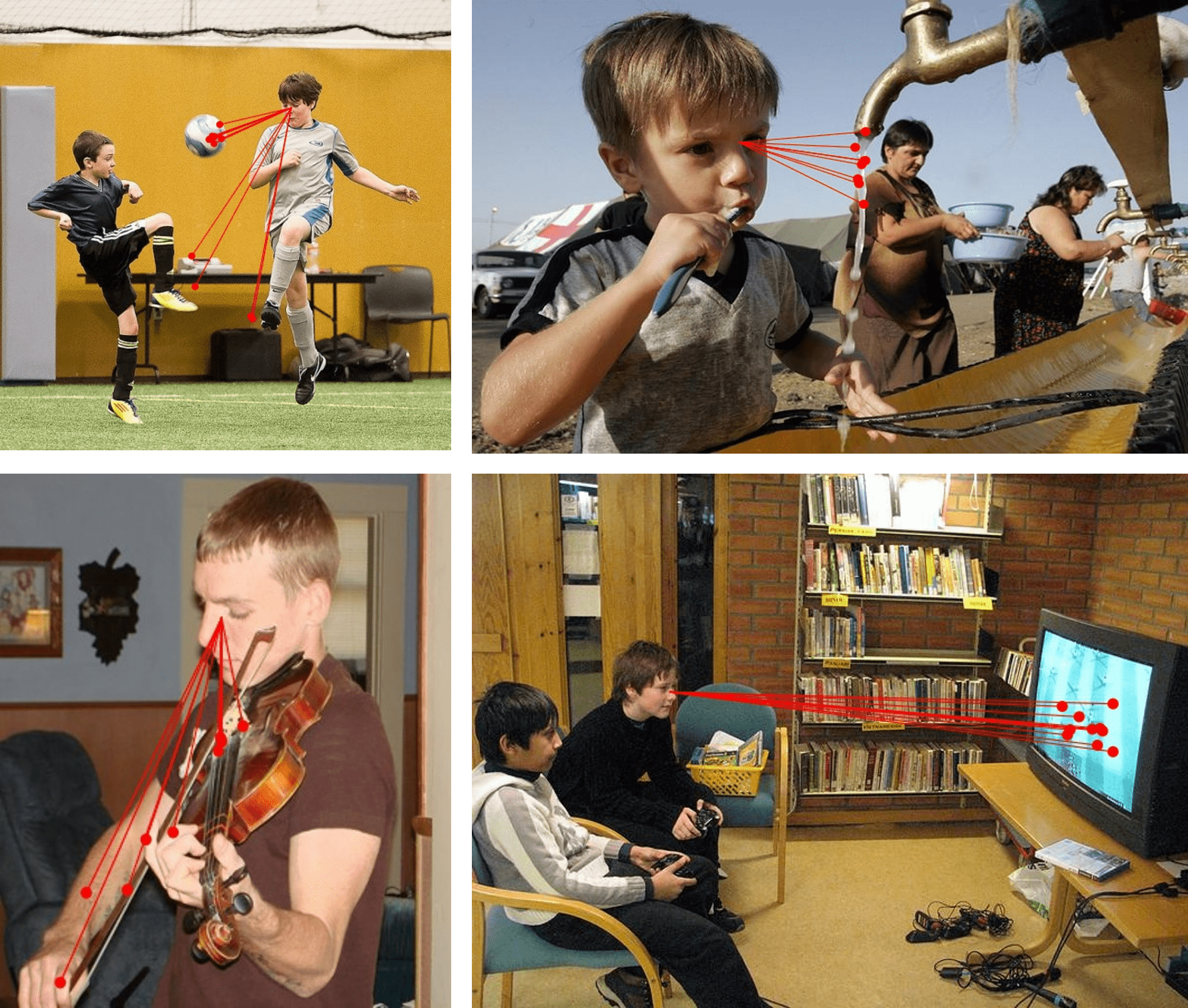}
    \caption{}
\end{subfigure}
\hspace{0.03\linewidth}
\begin{subfigure}{0.47\linewidth}
    \centering
    \includegraphics[width=\linewidth, height=1.15in]{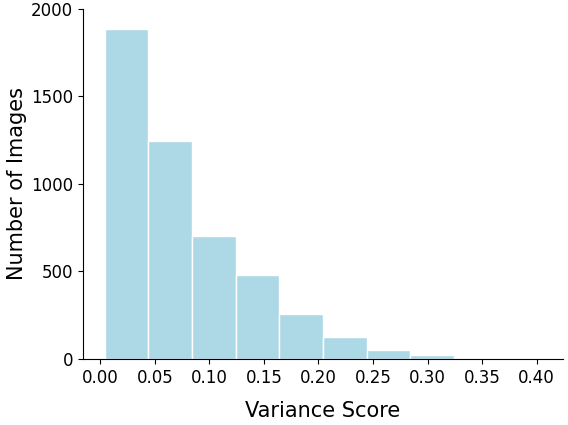}
    \caption{}
\end{subfigure}
\caption{Variance of gaze target annotations. (a) Annotations from multiple annotators on example images in the GazeFollow\cite{NIPS2015_ec895663} test set. Each red dot is the annotation from one annotator. Note that annotators often disagree on the exact gaze target. (b) The distribution of annotation variance for images in the GazeFollow test set. Please refer to Sec. 4.4.2 for the calculation of the variance score. }
\label{fig:gazefollow_info}
\end{figure}

Recent gaze following work formulated the target detection task as a heatmap prediction task \cite{lian2018believe,chong2020detecting,fang2021dual, Tu_2022_CVPR, Bao_2022_CVPR}. The heatmap prediction module is typically trained with a mean square error (MSE) loss with the ground truth heatmap, which is generated by applying a Gaussian kernel around the gaze target pixel. However, current gaze following datasets only provide one annotated coordinate for each person in the training set, while as in Figure \ref{fig:gazefollow_info}, the gaze target is usually ambiguous as different annotators may disagree on the exact location of the gaze target. Therefore, always requiring the model to regress to a unimodal Gaussian distribution in a higher resolution is not only a strict constraint in regression, but also biases the model towards predicting the same  distribution pattern  during inference, which lacks the consideration of annotator disagreement. Therefore, a regularization method is needed to relax the stricter constraint of the pixel-wise MSE loss, and should also encourage the model to predict more general distributions instead of a single Gaussian for uncertain images. 

In addition, besides estimating the target location, an effective gaze following model should also be capable of indicating if the target is located outside the image (in/out prediction). Previous work only trained the model with MSE loss and binary cross entropy (BCE) loss on the heatmap and in/out probability score predicted from two heads for the target prediction and in/out prediction subtasks \cite{{chong2018connecting,chong2020detecting,fang2021dual}}, without considering any correlation between them. However, we claim that these two subtasks should not be considered separately. Specifically, the outside case should be regarded as a special case of the target prediction task, except that the target is outside of the camera field-of-view. When a human follows someone's gaze in the image, he/she would directly perceive the location the person is looking at inside the image, or infer the person is looking at an unknown target outside the image, instead of considering the target and in/out prediction tasks separately. Therefore, we decide to model the gaze following behavior as a distribution of potential gaze targets over all possible locations, including an `outside' target. This enables us to consider the two subtasks in a holistic manner, which better mimics human gaze following behavior, and brings significant improvement in the in/out prediction task.

\begin{figure}[t]
\centering
    \includegraphics[width=\linewidth]{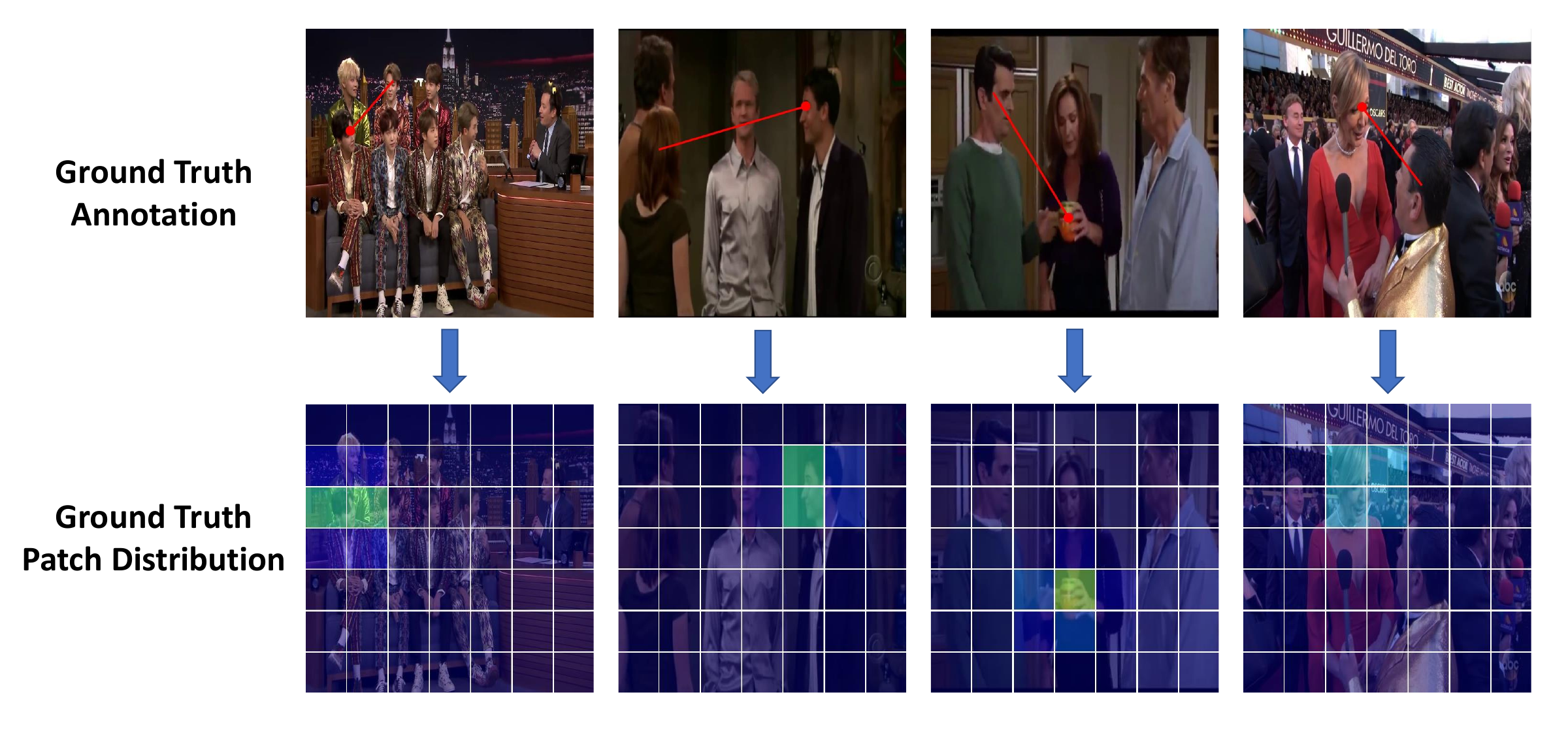}
\caption{Visualizations of the ground truth patch distributions created from different images. Note that our method can create various patterns of distributions, compared to the Gaussian ground truth which is always circular.}
\label{fig:patch_dist}
\end{figure}

In this paper, we propose the patch distribution prediction (PDP) for gaze following by replacing the in/out prediction task with the prediction of patch-level gaze distribution, which serves as a regularization for gaze heatmap prediction. Our PDP method regularizes the heatmap prediction from two perspectives. First, due to the coarser scale of patches, PDP has a softer constraint, which relaxes the pixel-wise MSE loss in higher resolution; In addition, as shown in Figure \ref{fig:patch_dist}, our patch distribution (PD) has variable patterns for different images with our creation method. As the feature tokens are associated with the patch responses one-to-one, the variable distribution pattern and high responses in multiple patches will enhance the generality of the common feature tokens, and encourage the heatmap prediction head to predict multi-modal heatmaps in the coarser scale instead of a single Gaussian for more uncertain images. Furthermore, PDP also bridges the gap between the target prediction and in/out prediction subtasks by predicting gaze distribution. With the introduction of an `outside' token, the gaze distribution can be predicted regardless of whether the target is located inside the image. Our claims are supported by our experiments.

Our main contributions can be summarized as follows:
\begin{itemize}
    \item We propose the PDP method for gaze following. To the best of our knowledge, we are the first to address the imperfections of gaze following training methods by considering multiple scales, generalizing the target distribution patterns and the correlations between the in/out prediction and gaze target prediction subtasks.
    \item Our model is especially effective on images with larger variance in annotations. By predicting heatmaps that are more aligned with group-level human annotations, our model can achieve a much higher, super-human Area Under Curve (AUC) on the GazeFollow dataset.
    \item Our model also shows a significant improvement in the in/out prediction task. By integrating the target prediction and in/out prediction subtasks with PDP, our model enables training with two tasks simultaneously without loss in performance in either task. 
    
\end{itemize}

\section{Related Work} \label{rw}

\textbf{Gaze Following.} GazeFollow \cite{NIPS2015_ec895663} is the first work focusing on unconstrained gaze target prediction with a deep learning model. Its two-branch design, with one branch encoding the scene saliency information, and the other encoding the head gaze information has been commonly adopted in later works \cite{chong2018connecting,lian2018believe,chong2020detecting,jin2021multi}. Later, Recasens \emph{et al.} \cite{recasens2017following} predicted the gaze target across temporal frames in videos. Chong \emph{et al.} \cite{chong2018connecting} considered the cases of the target located outside and trained the model in a multi-task learning approach. All these earlier methods formulated the gaze target prediction task as a one-hot patch classification task, which suffered from higher errors in distance metrics due to the coarse scale of patches.

\begin{figure*}[htbp]
    \centering
    \includegraphics[width=0.95\textwidth,height=6.5cm]{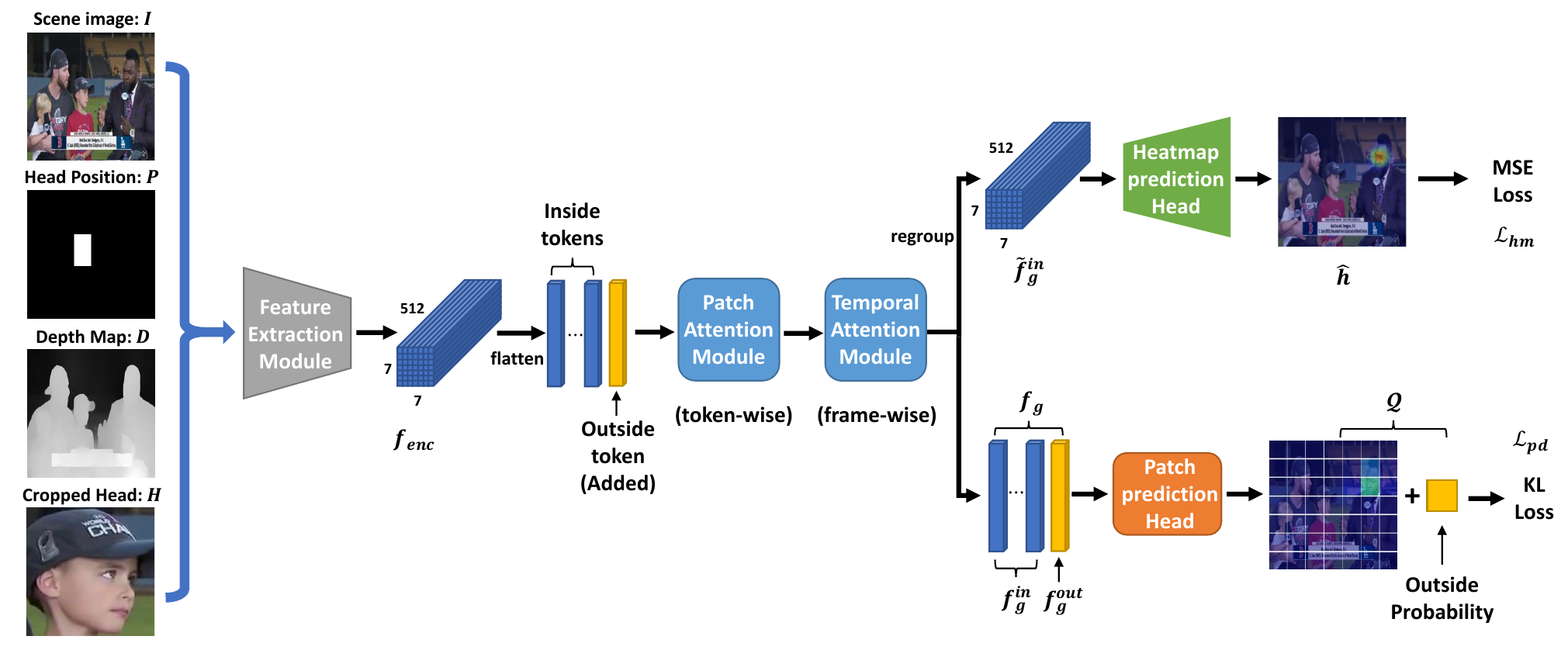}
    \caption{Overall structure of our model. The feature extraction module extracts feature encoding $f_{enc}$ from the input, which is then flattened in the spatial dimension and concatenated with an added `outside token'. The tokens go through the patch attention and temporal attention modules for information aggregation. Finally, the `inside tokens' are regrouped in the spatial dimension for heatmap prediction, while all tokens go into another patch prediction head for PDP.}
    \label{fig:whole_model}
\end{figure*}

Lian \emph{et al} \cite{lian2018believe} was the first work that formulated gaze target prediction as a heatmap regression task, using the scene image and the multi-scale gaze direction fields. Chong \emph{et al.} \cite{chong2020detecting} proposed the VideoAtt model and extended the task to Video Gaze Following. Later, Fang \emph{et al.} proposed the DualAtt model \cite{fang2021dual} by incorporating depth information, and 3D gaze direction estimated with eye images. Most recently, Tu \emph{et al.} proposed a transformer model for gaze following \cite{Tu_2022_CVPR}, and Bao \emph{et al} reconstructed the 3D scenes using depth maps and estimated human poses \cite{Bao_2022_CVPR}. All these models trained the heatmap regression task with MSE, except for Lian et al \cite{lian2018believe} which uses binary cross entropy (BCE) loss. Despite some level of uncertainty consideration with Gaussian smoothing, the constraint to always regress to a circular Gaussian in a pixel-wise manner limits the model's generality for predicting distribution in uncertain cases. Furthermore, all previous models treated the in/our prediction task as a binary classification task with a separate head, while none has considered correlation between the two subtasks.

Perhaps the most similar work to ours is \cite{zhang2018coarse}, which does a one-hot patch-level classification and a heatmap regression task for gaze prediction in egocentric videos. However, even if this method is optimal for egocentric gaze prediction, it is not optimal for the gaze following task due to the following reasons: The one-hot classification in the patch level is still unimodal with a fixed distribution pattern; The one-hot patch formulation may not represent the gaze distribution well if the target is close to the patch boundaries; The model cannot handle the case when the target is located outside. Furthermore, \cite{zhang2018coarse} shows cases of inconsistency between the patch classification and gaze prediction results, while our patch and heatmap predictions are highly consistent. We provide more detailed comparisons with the one-hot design in the supplementary material. %, but we want to note that \cite{zhang2018coarse} adopts a coarse-level patch design with a different model structure, which may still be suitable for their respective task domain.

\textbf{Gaze Object Prediction} detects the gazed object or objects of interest that may attract gaze in a scene. Massé et al \cite{masse2019extended} predict the positions of all objects of interest that may be out-of-frame in a top-down view heatmap by inferring the gaze directions of people in a video, but requires multiple people in the scene and the object positions to be known in the top-down view during training. Recently, the GOO dataset \cite{tomas2021goo} was released for gaze object prediction in retail environments, and a recent model was trained and evaluated on this dataset by doing object detection and target prediction with a shared backbone \cite{wang2022gatector}. However, the GOO dataset contains no outside case and mostly synthetic images, and the ground truths are pre-determined and not annotated by human annotators.

\textbf{3D Gaze Estimation} predicts a 3D gaze direction instead of the gaze target. Methods of 3D gaze estimation can be divided into model-based \cite{nakazawa2012point,shih2004novel,guestrin2006general,zhu2005eye} and appearance-based approaches \cite{zhang2015appearance,fischer2018rt,cheng2018appearance,krafka2016eye,park2019few,cheng2018appearance,kellnhofer2019gaze360}. Model-based approaches estimate the gaze direction from geometric eye features and models. Appearance-based approaches estimate gaze direction directly from face or eye images. Some recent 3D gaze estimation methods consider the uncertainty caused by the varying levels of difficulties of the input. Kellnhofer \emph{et al.} \cite{kellnhofer2019gaze360} used the pinball loss function \cite{koenker2001quantile} to compute the variance to a certain quantile for the gaze yaw and pitch angles. Dias \emph{et al.} \cite{dias2020gaze} used Bayesian neural networks to predict an uncertainty score with the input. However, these methods are difficult to be directly incorporated into gaze following tasks, as gaze following focus on target prediction instead of direction estimation, and the cases of multiple potential locations may make a single direction prediction suboptimal.
\section{Method} \label{method}
\subsection{Overall Structure}
As shown in Figure \ref{fig:whole_model}, our model consists of three components: feature extraction, gaze distribution feature computation, and two heads for heatmap and patch-level gaze distribution prediction. The feature extraction module takes in the scene image $I \in R^{3 \times H_0 \times W_0}$, the binary head position mask $P  \in \{0, 1\}^{H_0 \times W_0}$, a normalized depth map $D \in [0,1]^{H_0 \times W_0}$, and the cropped head of the person $H \in R^{3 \times H_0 \times W_0}$ as input, and outputs the extracted feature $f_{enc} \in R^{C \times H \times W}$. 
The design of the feature extraction module generally follows the VideoAtt model \cite{chong2020detecting}, except that we leveraged an additional depth map as input according to the insight from the DualAtt model \cite{fang2021dual} to incorporate scene depth information, with some small modifications. Please refer to the supplementary material for details of the feature extraction module.

Subsequently, the gaze distribution feature computation component operates on $f_{enc}$ and outputs the gaze distribution feature $f_{g} \in R^{C \times (H \cdot W+1)}$, which consists of the inside tokens $f_{g}^{in} \in R^{C \times (H \cdot W)}$  and an outside token $f_{g}^{out} \in R^{C}$. Finally, $f_{g}^{in}$ is regrouped and fed into the heatmap prediction head to output the heatmap $\hat{h} \in R^{C \times H' \times W'}$  , while all tokens go into the patch prediction head, and output the patch-level gaze distribution $\mathcal{Q}=\{q_i\}_{i=1}^{H \cdot W+1}$. We will illustrate each component in detail below. 

\subsection{Gaze Distribution Feature Computation}
\begin{figure}[t]
\centering
    \begin{subfigure}{0.45\linewidth}
    \centering
    \includegraphics[width=0.95\linewidth, height=1.85in]{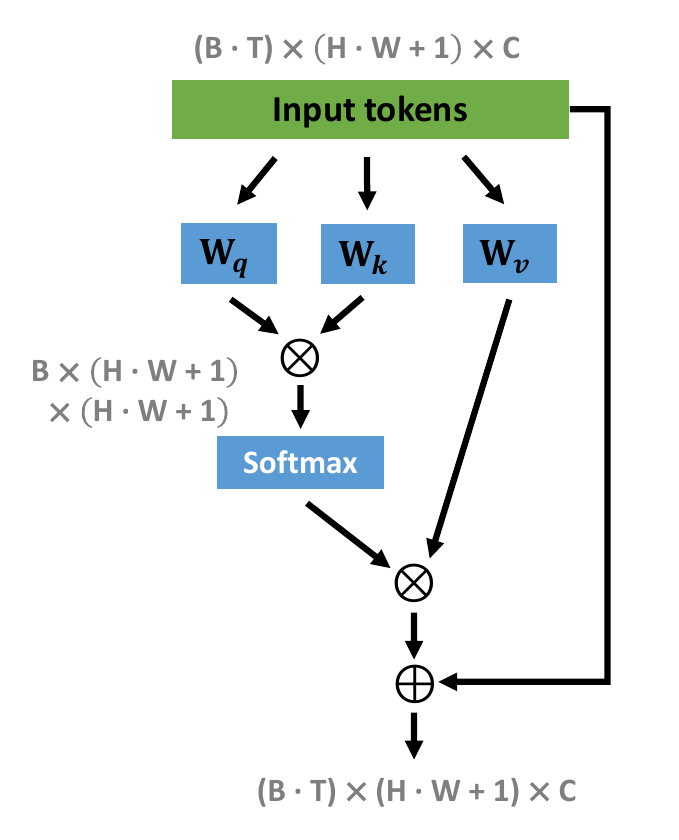}
    \caption{}
\end{subfigure}
\hspace{0.05\linewidth}
\begin{subfigure}{0.45\linewidth}
    \centering
    \includegraphics[width=\linewidth, height=1.85in]{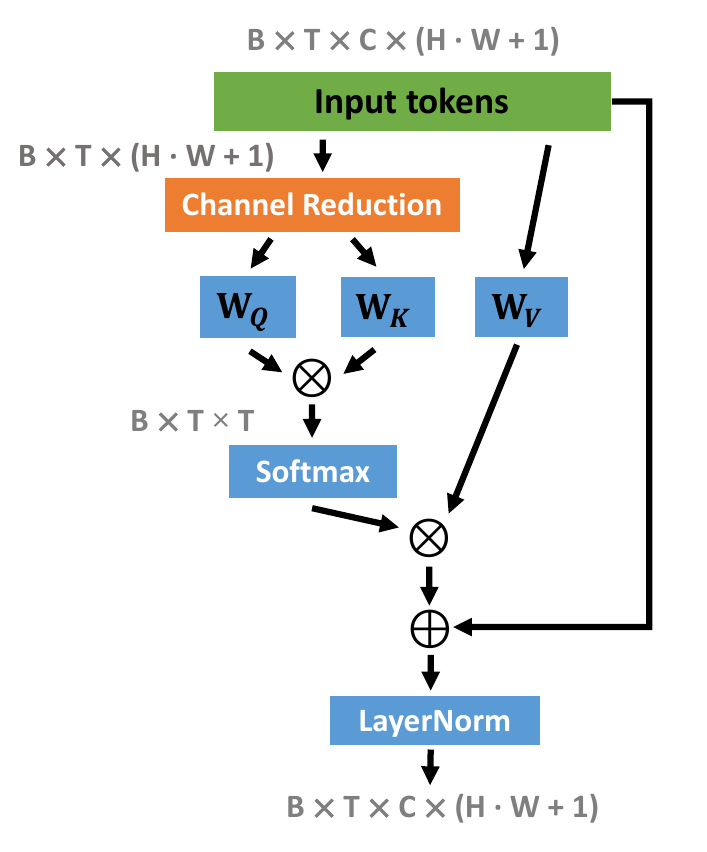}
    \caption{}
\end{subfigure}
\caption{Details of the (a) patch attention module and (b) temporal attention module.}
\label{fig:att_modules}
\end{figure}

This component is responsible for getting the gaze distribution feature before PDP and heatmap prediction. The output $f_{enc}$ from the feature extraction module has a shape of $C \times H \times W$. We can consider that each $C$ dimensional feature vector represent one patch in the image after dividing the image in to $H \times W$ patches. Therefore, $f_{enc}$ can be regarded as $H \cdot W$ 
`inside tokens' with C dimensions. $f_{enc}$ is then flattened to the shape of $C \times (H \cdot W)$. To obtain the gaze distribution in both inside and outside cases, we add an `outside token' $x_{out} \in R^C$, and concatenate it with the flattened and transposed version of $f_{enc}$ and get $f_{enc}' \in R^{(H \cdot W +1) \times C}$. The `outside token' is a learnable parameter vector that is trained along with the model. $f_{enc}$ is then fed into the patch attention and temporal attention modules (Figure \ref{fig:att_modules}) to get the gaze distribution feature $f_{g}$.

The patch attention module $PA(\cdot)$ is introduced to make each patch token have a better understanding of the global scene information before computing the overall gaze distribution. It is a dot-product self attention module similar to \cite{ramachandran2019stand}, but we added learnable positional embeddings to all tokens to make the attention module aware of the positions of the tokens. In $PA(\cdot)$, each token in one spatial location is added with a weighted sum of all tokens, where the weights are computed from the dot product of the feature vectors in a pairwise manner and normalized by a softmax function:
\begin{equation}
    \hat{f}_{g} = f_{enc}' + \mathrm{softmax}(q k^\top)v ,
\end{equation}
where $ q = W_q f_{enc}'$, $k = W_k f_{enc}'$, $v = W_v f_{enc}'$ are the queries, keys and values respectively. $W_q, W_k, W_v \in R^{C \times C}$ are learnable linear projections. 

In video gaze following, the gaze distribution feature in nearby time steps should be helpful for the prediction of the current time step. To aggregate the information in the temporal dimension, we introduce the temporal attention module $TA(\cdot)$. The structure of $TA(\cdot)$ is similar to the patch attention module, but we made some modifications to make it work more efficiently. Suppose we have a sequence of output features $\hat{F}_{g} = \{ \hat{f}_{gi} \}_{i=1}^{T}$ from a sequence of input images $\mathcal{I} = \{ I_i\}_{i=1}^T$ in the video after the patch attention module. In order to reduce memory and computational load for calculating attention between frames, a convolutional layer is used to map $\hat{F}_{g}$ to 1 single channel and get $\hat{F}_{g}' \in R^{T \times (H \cdot W + 1)}$ (we also tested mapping to more channels but did not observe improvement). Temporal attention weights are computed similarly as $PA(\cdot)$: $M_{att} = \mathrm{softmax}(QK^\top)$, except the queries, keys and values are computed from the compressed feature map $Q = W_Q \hat{F}_{g}'$, $K = W_K\hat{F}_{g}'$, $V = W_V\hat{F}_{g}'$, $W_Q, W_K, W_V \in R^{(H \cdot W + 1) \times (H \cdot W + 1)}$. Finally, the original input features $\hat{F}_{g}$ are aggregated with the attention-weighted values with a residual connection to get the gaze distribution feature $F_g= \{f_{gi}\}_{i=1}^{T}$. We found applying a LayerNorm on the output can lead to more stable training:
\begin{equation}
    F_{g} = LayerNorm(\hat{F}_{g}+ f(M_{att}V)),
\end{equation}
The temporal attention module will be removed for inference on a single image. In this case, the gaze distribution feature will be the output of $PA(\cdot)$: $f_{g} = \hat{f}_{g}$.

\subsection{Gaze Prediction}
The gaze distribution feature $f_g$ is finally fed into two heads for gaze heatmap regression and PDP. The heatmap prediction head consists of one convolutional layer, followed by 3 deconvolutional layers, and a final convolutional layer, which is similar in structure to the ones in the VideoAtt \cite{chong2020detecting} and DualAtt \cite{fang2021dual} model, but we replaced their in/out prediction head with our patch distribution prediction head, which consists of two fully connected layers operating on the channel dimension to get the patch probability score for each token: 
\begin{equation}
    \pi_q = \sigma(h_2(Relu(h_1(f_g)))),
\end{equation}
where $\sigma$ indicates sigmoid function. The PD can be obtained by normalizing each token's score:
\begin{equation}
    q_{i} = q(g_i = 1|X) =  \frac{\pi_q^{i}}{\sum_{j=1}^{H \cdot W +1}\pi_q^{j}},
\end{equation}

where $q_{i}$ is the probability confidence of the gaze target locating in token $i$, $X$ indicates all inputs to the model. Therefore, the probability that the target is located inside the image is the sum of scores from all inside tokens: $P_{in} = \sum_{j=1}^{H \cdot W }q_{j}$. For gaze heatmap regression, the $H \cdot W$ `inside tokens' $f_g^{in}$ are regrouped to a spatial feature map $\widetilde{f}_g^{in} \in R^{C' \times H \times W}$. $\widetilde{f}_g^{in}$ is then fed into the heatmap prediction module to generate the output gaze heatmap  $\hat{h}$. 

In training, the heatmap regression loss $\mathcal{L}_{hm}$  is the MSE loss between the predicted and ground truth heatmap. KL-divergence loss is chosen for the PDP task with the predicted and ground truth PD:
\begin{equation}
    \mathcal{L}_{pd} = KL(q(g|X)||p(g|X))
\end{equation}
The final loss is a weighted sum of the two losses:
\begin{equation}
    L = \lambda_1 \cdot \mathcal{L}_{hm} + \lambda_2 \cdot \mathcal{L}_{pd}
\end{equation}

\subsection{Ground Truth Patch Distribution Creation}
\begin{figure}[t]
\centering
    \includegraphics[width=\linewidth]{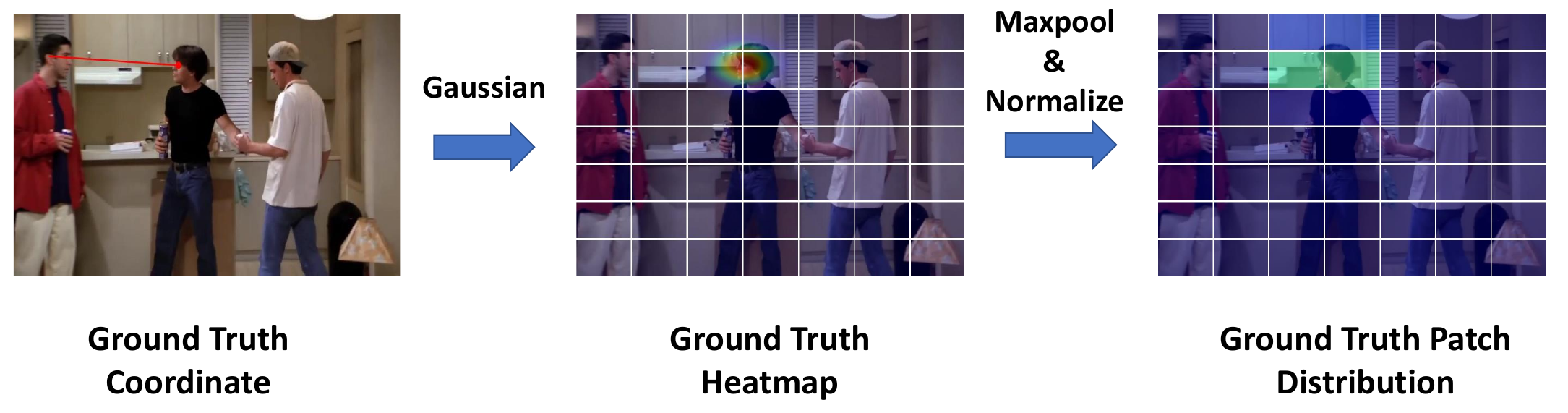}
\caption{Procedure of Ground truth PD creation.}
\label{fig:gt_patch}
\end{figure}

In order to train the subtask of PDP, a ground truth patch-level gaze distribution is generated, of which the procedure is shown in Figure \ref{fig:gt_patch}.

The ground truth patch distribution is created from the ground truth heatmap for the heatmap prediction task, which is generated by applying a gaussian kernel around the annotated gaze coordinate:
\begin{equation}
    h(j,k) = \frac{1}{\sqrt{2\pi}\sigma}\exp{-\frac{(j-g_x)^2+(k-g_y)^2}{2\sigma^2}}
\end{equation}

\noindent where $g=(g_x, g_y)$ is the annotated gaze coordinate. 

For each patch, the points in the heatmap within that patch are located, and the maximum heatmap score is taken from these points as the probability score of that patch:
\begin{equation}
    \pi_p^{i} = \max_{(j,k) \in \mathcal{N}(i)} (h(j,k)),
\end{equation}

\noindent where $\mathcal{N}(i)$ is the pixels in the heatmap that are within patch $i$. We obtain the patch-level distribution value $ \widetilde{\pi}_p^{i} $ by dividing  $\pi_p^{i}$ by the summed scores from all patches. If target is located outside, the outside token will have a probability of 1 and all the inside tokens will have a probability of 0:
\begin{equation}
     p_i = 
    \begin{cases}
     \widetilde{\pi}_p^{i} \quad if \; Y=1 \; else \; 0 & i \leqslant H \cdot W \\
     1 - Y, & i = H \cdot W + 1 \\
    \end{cases},
\end{equation}
$Y=1$ if the target is located in the image, otherwise $Y=0$.

With this discretization and normalization method, we can obtain various distribution patterns for the ground truth patch distribution, usually with high responses in multiple patches, as shown in Figure \ref{fig:patch_dist}. By encouraging the model to predict higher responses in different patches, the model will tend to predict multi-modal heatmap clusters in the coarser scale on images with ambiguous targets, with a shared feature embedding before the two prediction heads. 
We provide the implementation details and results of other alternatives of the patch distribution creation settings in the supplementary material.

\section{Experiments and Results} \label{ex_and_res}
\subsection{Datasets}
\textbf{GazeFollow} \cite{NIPS2015_ec895663} is a large-scale image dataset for gaze following. The dataset contains over 122K images in total with over 130K people inside the images, of which 4782 people were used for testing. For human consistency, 10 annotations were collected per person in the test set, while the training set just contain 1 annotation per person. Later, Chong \emph{et al.} \cite{chong2018connecting} extended it with more accurate annotations of whether the gaze target is located inside the image. \textbf{VideoAttentionTarget}  \cite{chong2020detecting} is a dataset for gaze following in videos. Video clips were collected from 50 different shows on Youtube, each of which has a length between 1-80 seconds. The dataset consists of 1331 head tracks with 164K frame-level bounding boxes, 109,574 in-frame gaze targets, and 54,967 out-of-frame gaze indicators. Both the training and test sets contain only 1 annotation per person. 

\subsection{Evaluation Metrics}
\textbf{Area Under Curve (AUC)} is commonly adopted by the gaze following works \cite{NIPS2015_ec895663,chong2018connecting,lian2018believe,chong2020detecting,fang2021dual} for assessing the confidence of the predicted heatmap. The ground truth in GazeFollow is the annotations from all 10 annotators, while in VideoAttentionTarget is the heatmap created from the single annotated coordinate. \textbf{Dist}: $L^2$ distance between annotated gaze coordinate and predicted location, determined as the point with maximum confidence on the heatmap. Specifically, in GazeFollow dataset, both \textbf{Min Dist.} and \textbf{Avg Dist.} are calculated. \textbf{In/Out AP}: Average Precision (AP) is used in the evaluation of In/Out prediction based on predicted probability of the gaze target locating in frame.

\subsection{Results}
\begin{table}[t]
    \centering
    \begin{tabular}{l c c c c}
        \toprule
       \multirow{2}{*}{Method} & \multirow{2}{*}{Dep.} & \multirow{2}{*}{ AUC $\uparrow$} & \multicolumn{2}{c}{ Dist. $\downarrow$}  \\
                        &                       &                      & Avg. & Min. \\ 
        \toprule
        Random \cite{NIPS2015_ec895663} & $\times$ & 0.504 & 0.484 & 0.391 \\
        Center \cite{NIPS2015_ec895663} &  $\times$ & 0.633 & 0.313 & 0.230 \\
        Fixed Bias \cite{NIPS2015_ec895663} & $\times$ & 0.674 & 0.306 & 0.219 \\
        GazeFollow \cite{NIPS2015_ec895663} & $\times$ & 0.878 & 0.190 & 0.113 \\
        Chong \emph{et al.} \cite{chong2018connecting} & $\times$ & 0.896 & 0.187 & 0.112 \\
        Lian \emph{et al.} \cite{lian2018believe} & $\times$ & 0.906 & 0.145 & 0.081 \\
        VideoAtt \cite{chong2020detecting} & $\times$ & 0.921 & 0.137 & 0.077 \\
        VideoAtt\_depth & $\checkmark$ & 0.927 & 0.131 & 0.071 \\
        DualAtt \cite{fang2021dual} & $\checkmark$ & 0.922 & \underline{0.124} & \underline{0.067} \\
        HGTTR \cite{Tu_2022_CVPR} & $\times$ & 0.917 & 0.133 & 0.069 \\
        ESCNet \cite{Bao_2022_CVPR} & $\checkmark$ & \underline{0.928} & 0.126 & / \\
        \toprule
        Human & / & 0.924 & 0.096 & 0.040 \\
        \toprule
        Ours w.o. dep. & $\times$ & \underline{0.928} & 0.131 & 0.072 \\
        Ours & $\checkmark$ & \textbf{0.934}  & \textbf{0.123} & \textbf{0.065} \\
        \bottomrule
    \end{tabular}
    \caption{Evaluation of the spatial model part on GazeFollow dataset. Best numbers are marked as bold and 2nd best are underlined. Dep. indicate depth input.}
    \label{tab:gazefollow_results}
\end{table}

\begin{table}[t]
    \vspace{-.1cm}
    \centering
    \begin{tabular}{l c c c c}
        \toprule
        \multirow{2}{*}{Method} & \multirow{2}{*}{Dep.}& \multicolumn{2}{c}{\emph{In frame}}& \emph{Out of frame} \\ & &AUC $\uparrow$ & Dist $\downarrow$ & AP $\uparrow$\\ 
        \toprule
        Random \cite{chong2020detecting} & $\times$ & 0.505 & 0.458 & 0.621 \\
        Fixed Bias \cite{chong2020detecting} & $\times$ & 0.728 & 0.326 & 0.624 \\
        Chong \emph{et al.} \cite{chong2018connecting}  & $\times$ & 0.830 & 0.193 & 0.705 \\
        VideoAtt \cite{chong2020detecting} & $\times$  & 0.860 & 0.134 & 0.853 \\
        VideoAtt\_depth & $\checkmark$ & \underline{0.911}  & 0.118 & 0.861 \\
        DualAtt \cite{fang2021dual} & $\checkmark$ & 0.905 & \textbf{0.108} & \underline{0.896} \\
        HGTTR \cite{Tu_2022_CVPR} & $\times$ & 0.904 & 0.126 & 0.854 \\
        ESCNet \cite{Bao_2022_CVPR} & $\checkmark$ & 0.885 & 0.120 & 0.869 \\
        \toprule
        Human & / & 0.921 & 0.051 & 0.925 \\
        \toprule
        Ours w.o. dep. & $\times$ & 0.907 & 0.116 & 0.881 \\
        Ours & $\checkmark$ & \textbf{0.917} & \underline{0.109} & \textbf{0.908} \\
        \bottomrule
    \end{tabular}
    \caption{Evaluation of the full model on  VideoAttentionTarget dataset.}
    \label{tab:videoatt_results}
\end{table}

We evaluate the spatial part of our model (without temporal attention module) on the GazeFollow dataset \cite{NIPS2015_ec895663}. Table \ref{tab:gazefollow_results} shows our model's performance with the current state-of-the-art (SOTA) models. For a fair comparison with VideoAtt \cite{chong2020detecting}, we modified its feature extraction module as ours, and train with the scene depth map as additional input. We name this model as VideoAtt\_depth. It can be seen that our method outperforms the VideoAtt\_depth model by a large margin in all metrics, and also outperforms the other newer models \cite{fang2021dual, Tu_2022_CVPR, Bao_2022_CVPR} in the AUC metric. The AUC is the most important metric that evaluates the model's predicted heatmap with the group-level annotations, and our model can achieve super-human AUC even without depth input. The smaller performance advantage in the distance metrics may be because the PDP task emphasizes on predicting the gaze distribution instead of a specific gaze coordinate, which is the maximum point in the heatmap. Besides depth input, the DualAtt model \cite{fang2021dual} takes cropped eye images as input using head pose and face keypoint estimation models, and the ESCNet \cite{Bao_2022_CVPR} estimates human pose for 3D scene reconstruction. Our model does not require these additional inputs, and can achieve higher AUC than ESCNet  which claims to have multi-modal output. 

Table \ref{tab:videoatt_results} summarizes the performance of our full model on the VideoAttentionTarget \cite{chong2020detecting} dataset. Besides improvement in the target prediction metrics, our method shows a significantly higher in/out AP score compared to all SOTA models, closer to human performance. This supports our claim that the distribution prediction task bridges the gap between the target prediction and in/out prediction subtasks.   

\begin{figure*}[!t]
\setlength\tabcolsep{3pt}%%
\centering
\begin{tabular}{cccc}
    \textbf{Input} &
 \textbf{Patch prediction} &
 \textbf{Heatmap output} &
 \textbf{Target prediction}\\
 \includegraphics[width=1.5in, height=1.05in]{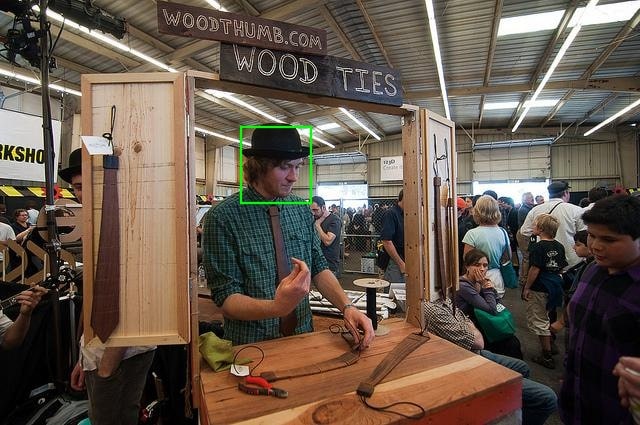} &
 \includegraphics[width=1.5in, height=1.05in]{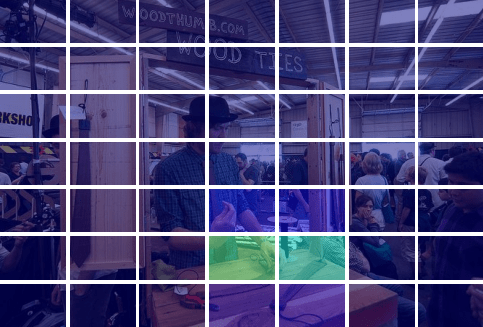} &
 \includegraphics[width=1.5in,height=1.05in]{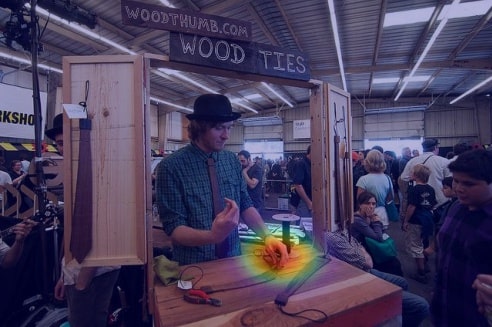} &
 \includegraphics[width=1.5in,height=1.05in]{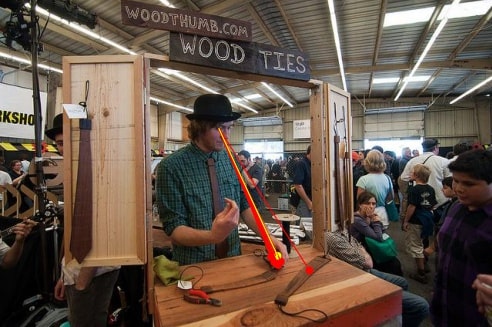}\\
 \includegraphics[width=1.5in, height=1.05in]{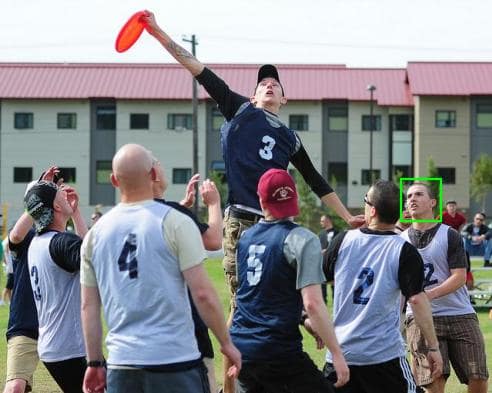} &
 \includegraphics[width=1.5in, height=1.05in]{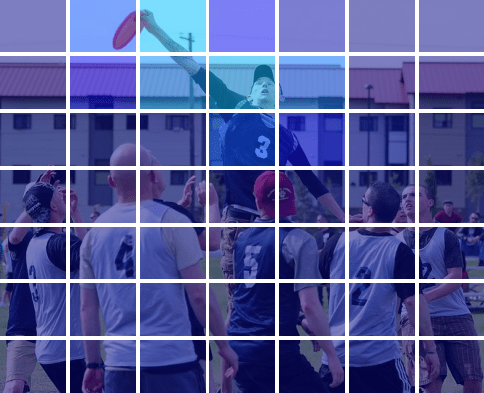} &
 \includegraphics[width=1.5in,height=1.05in]{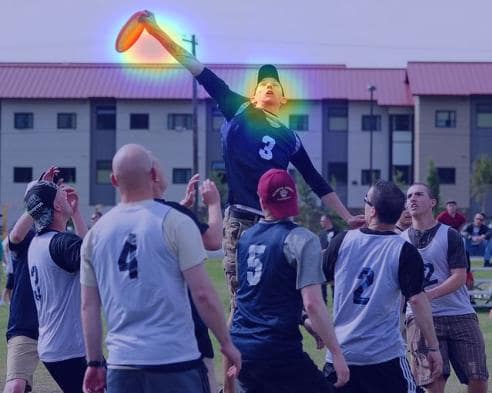} &
 \includegraphics[width=1.5in,height=1.05in]{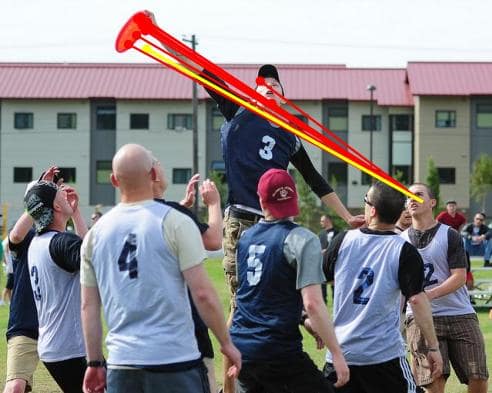}\\
 \includegraphics[width=1.5in, height=1.05in]{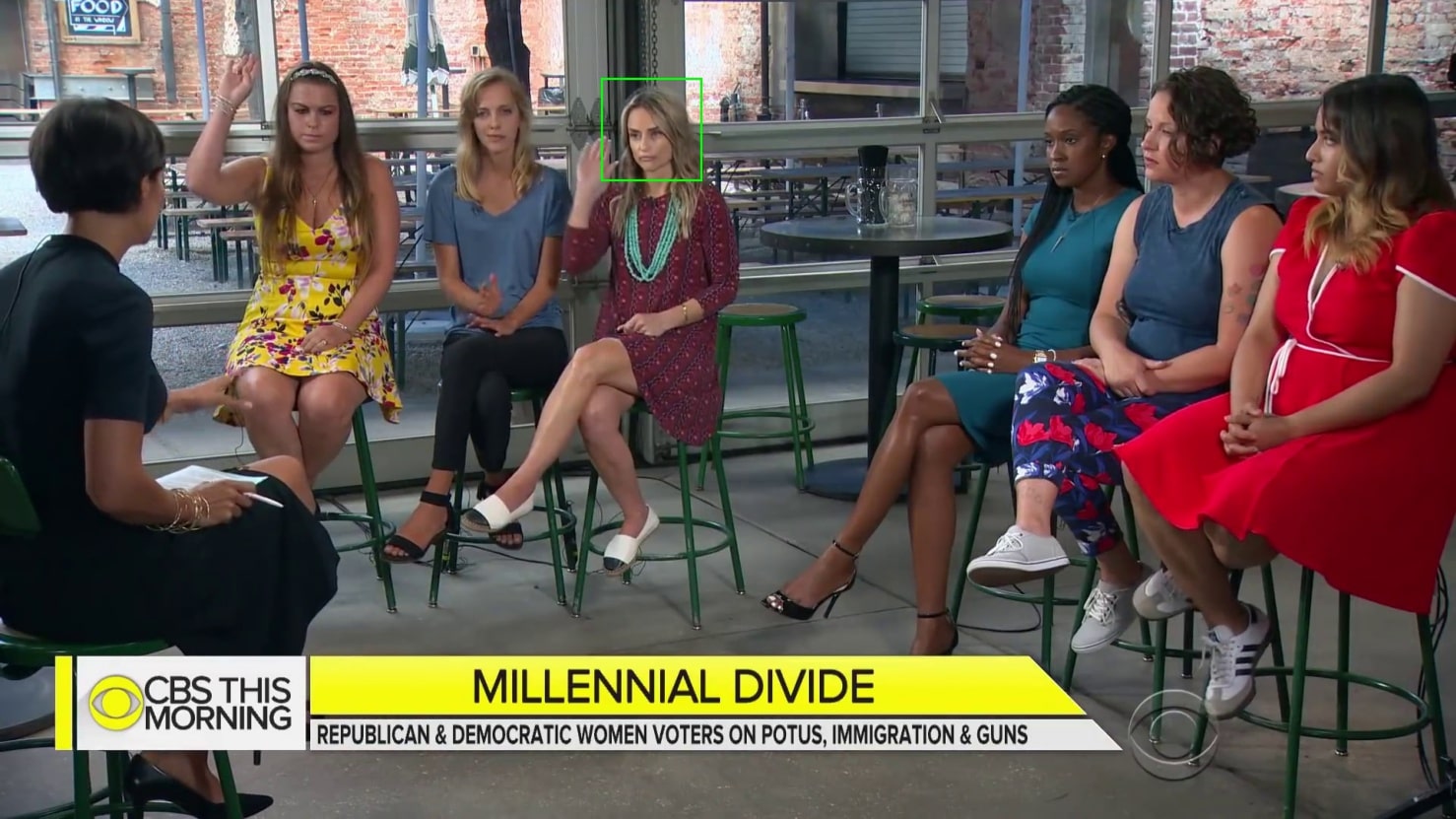} &
 \includegraphics[width=1.5in, height=1.05in]{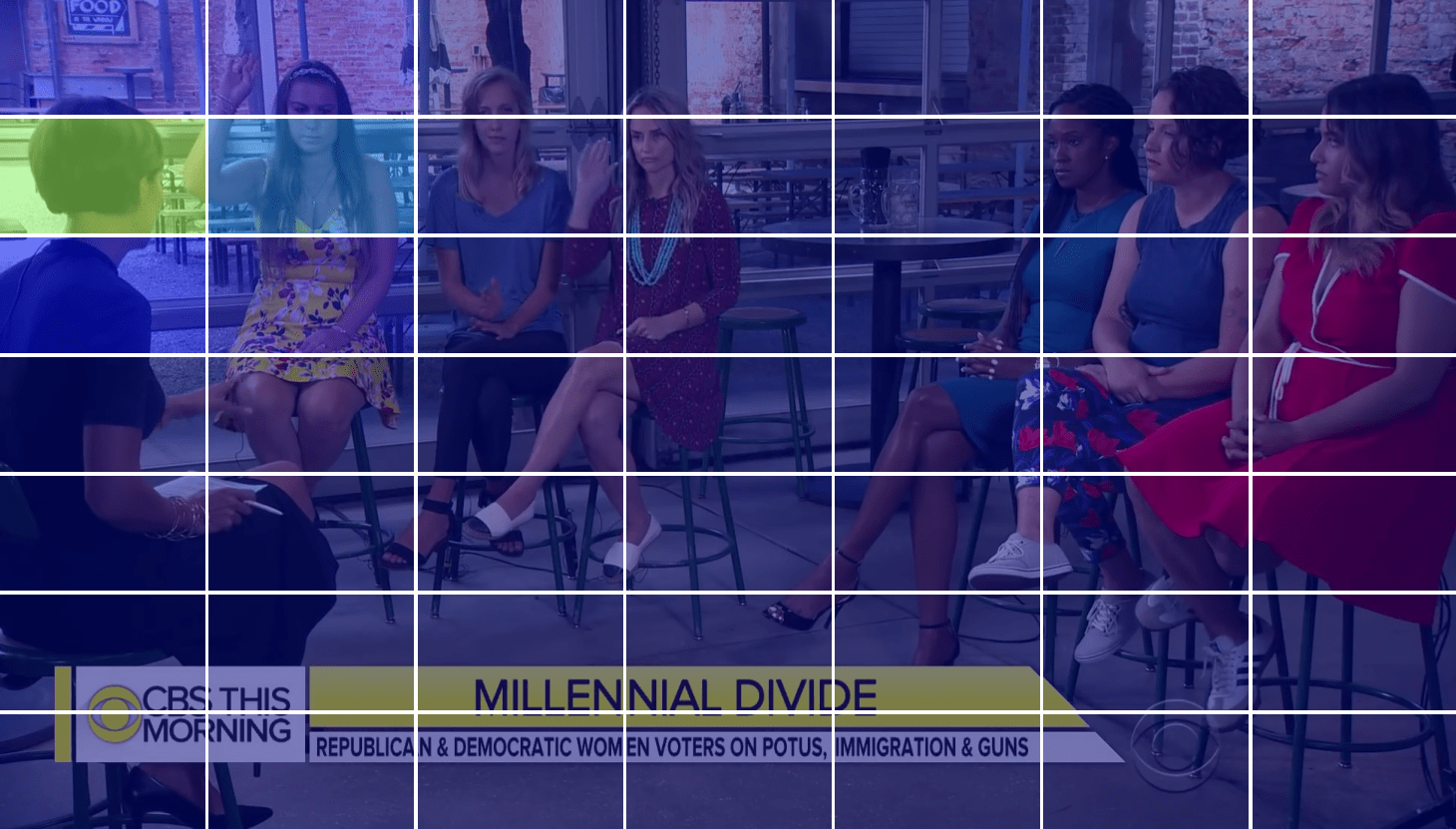} &
 \includegraphics[width=1.5in,height=1.05in]{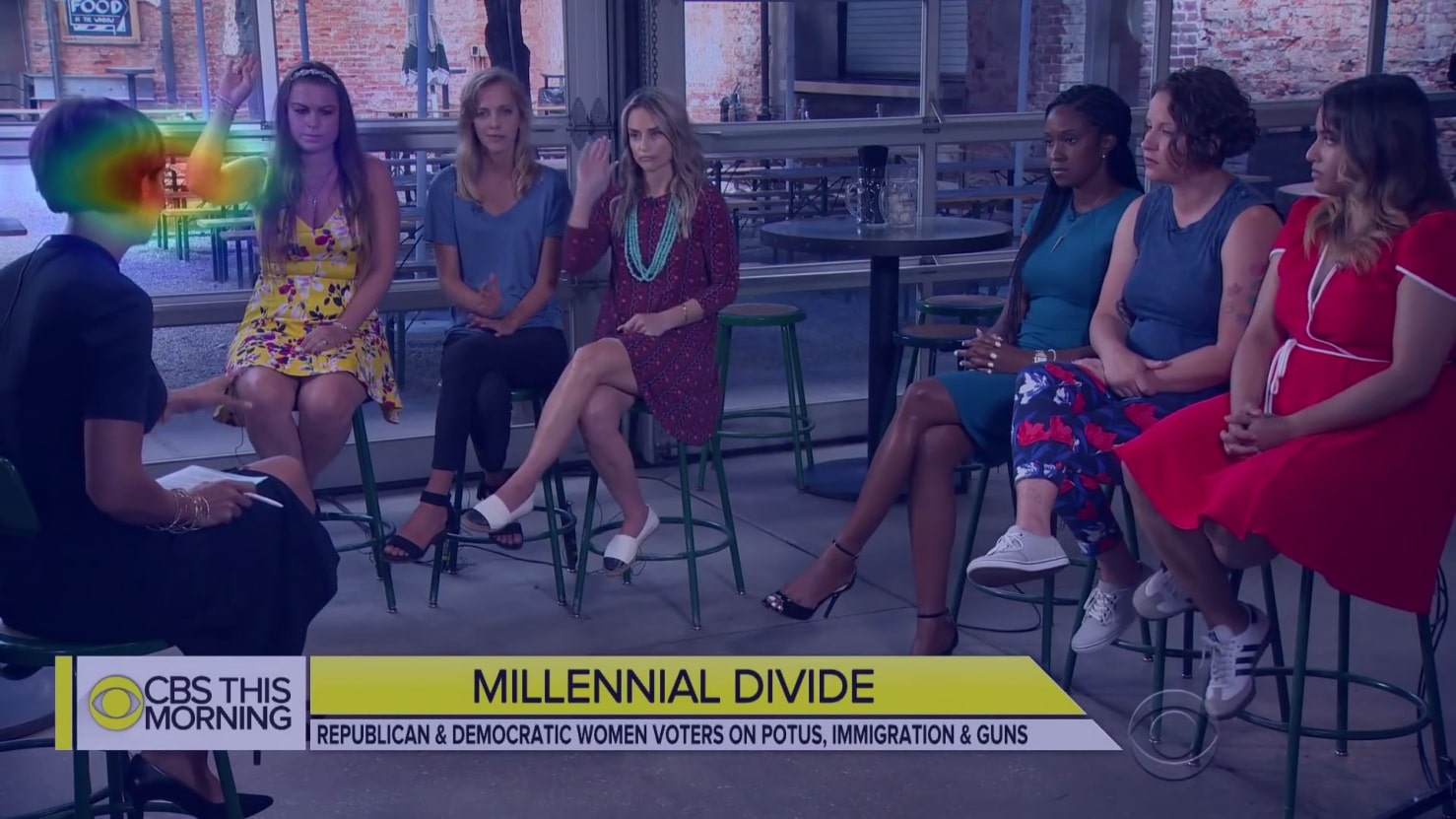} &
 \includegraphics[width=1.5in,height=1.05in]{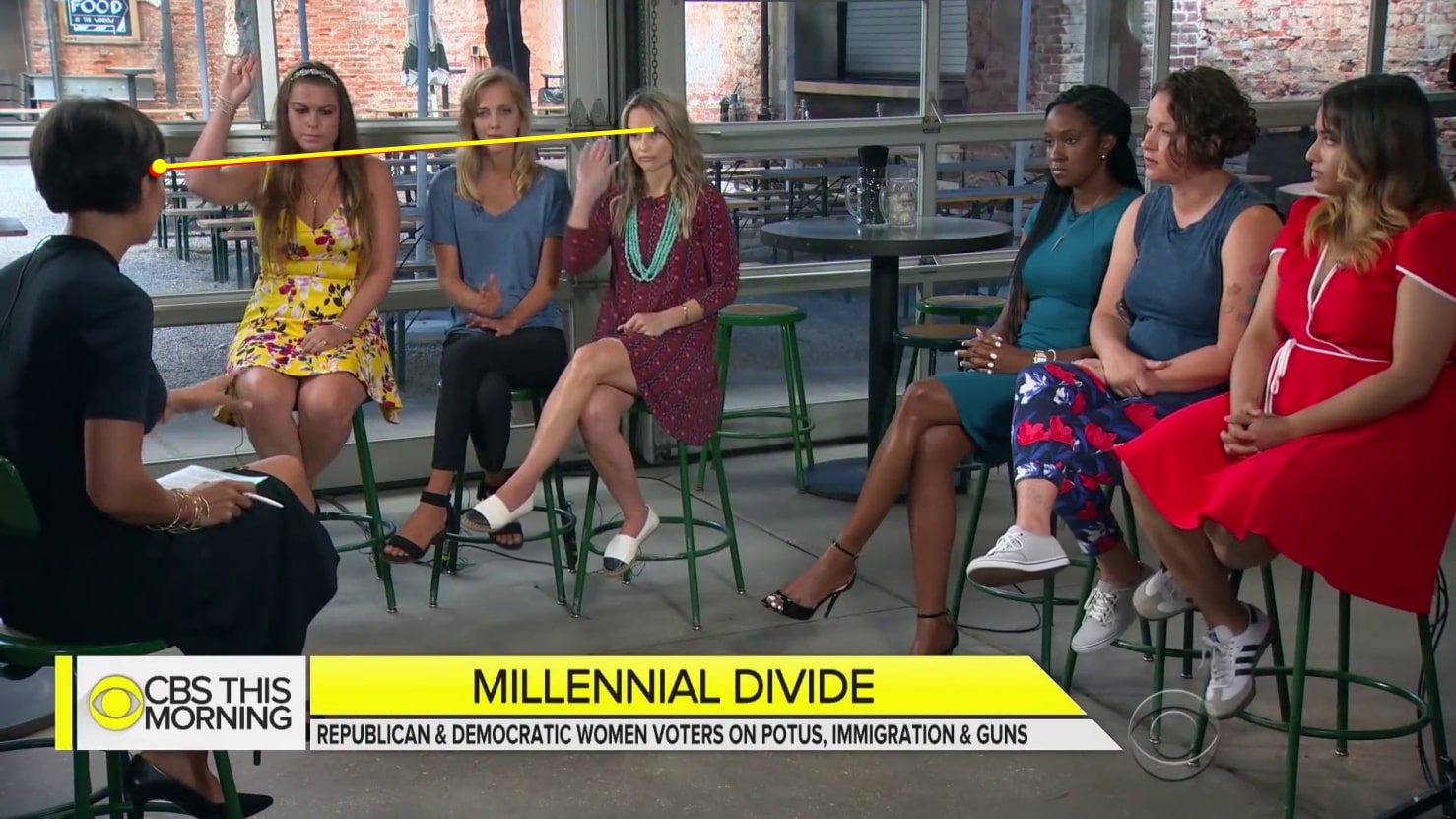}\\

\end{tabular}
\caption{Visualizations of output of our model on GazeFollow (1st and 2nd rows) and VideoAttentionTarget dataset (3rd row). Ground truth annoatations are visualized in red and the predicted point is visualized in yellow.}
\label{fig:qualitative_vis}
\end{figure*}

\begin{table*}[htbp]
\centering
\begin{tabular}{lcccccc}
\toprule
%\multirow{3}{*}{Method} &  \multicolumn{3}{c}{GazeFollow}                      & \multicolumn{3}{c}{VideoAttentionTarget}                                                                                                             \\ \cmidrule(lr){2-4} \cmidrule(lr){5-7}      
                       %& \multicolumn{1}{l}{\multirow{2}{*}{AUC $\uparrow$}} & \multicolumn{1}{l}{\multirow{2}{*}{Avg Dist. $\downarrow$}} & \multicolumn{1}{l}{\multirow{2}{*}{Min Dist. $\downarrow$}} & \multicolumn{2}{c}{\textit{In frame}} & \textit{Out of frame}  \\
                       %             &
                       %\multicolumn{1}{l}{}                     & \multicolumn{1}{l}{}                           & \multicolumn{1}{l}{}
                    %    & AUC $\uparrow$          & Dist. $\downarrow$              & AP $\uparrow$                                          \\ \toprule

\multirow{3}{*}{Method} & \multicolumn{3}{c}{GazeFollow}                   & \multicolumn{3}{c}{VideoAttentionTarget}                                                       \\ \cmidrule(lr){2-4} \cmidrule(lr){5-7}  
                        & \multirow{2}{*}{AUC $\uparrow$} & \multicolumn{2}{c}{Dist. $\downarrow$} & \multicolumn{2}{c}{\textit{In frame}}              & \multicolumn{1}{c}{\textit{Out of frame}} \\
                        &                      & Avg.        & Min.        & \multicolumn{1}{c}{AUC $\uparrow$} & \multicolumn{1}{c}{Dist $\downarrow$} & \multicolumn{1}{c}{AP $\uparrow$} \\ \toprule
No patch pred     & 0.927                                    & 0.138                                          & 0.077   & 0.889         & 0.130                 & 0.885                                                        \\
KL $\rightarrow$ MSE &  0.928 & 0.131     & 0.072 & 0.908 & 0.118 & 0.892  \\ 
KL $\rightarrow$ BCE &  0.931 & 0.127     & 0.067 & 0.904 & 0.116 & 0.890  \\ 
No patch att  &  0.929                                    & 0.125                                          & 0.066  & 0.902     & 0.119                 & 0.900                                                           \\
No temporal att          & /                                        & /                                              & /    & 0.914         & 0.111        & 0.906    \\
Add dir pred       & 0.931                & 0.124                                                    & \textbf{0.065}       & /             & /                     & /                                                              \\
\toprule
Full Model         & \textbf{0.934}                                    & \textbf{0.123}                                          & \textbf{0.065} & \textbf{0.917}         & \textbf{0.109}                 & \textbf{0.908}          \\
\bottomrule
\end{tabular}
\caption{Ablation Study Results}
\label{tab:ablation}
\end{table*}

Figure \ref{fig:qualitative_vis} shows the qualitative results. It can be seen that our model can predict both single and multi-modal heatmaps and patch distributions. There are good correspondences between the predicted patch distributions, the predicted heatmaps and the gazed objects.

\subsection{Analyses}
\subsubsection{Ablation Study}
Table \ref{tab:ablation} summarizes our ablation study results. The model still has a strong performance without the temporal attention module. However, there is an obvious drop in performance after removing the patch attention module on VideoAttenionTarget. This is understandable because, without the patch attention module, the tokens lose some global context, and the outside token will have a fixed prediction score for any input, making the distribution prediction difficult. We then tested replacing KL divergence with MSE or BCE which were commonly used in gaze heatmap prediction \cite{lian2018believe, chong2018connecting, fang2021dual, Tu_2022_CVPR, Bao_2022_CVPR}. Training with MSE showed a large drop in performance. BCE showed better performance than MSE due to higher robustness to noises, but is still worse than KL divergence. This demonstrates the importance of formulating PDP as a distribution prediction task. Finally, we replaced the PDP head with the original in/out prediction head in previous models \cite{chong2020detecting,fang2021dual}, by predicting an in/out probability score from $f_g$, and followed their training settings (no in/out loss for GazeFollow, and BCE for in/out loss in VideoAttentionTarget). The performance had a significant drop and became even worse than VideoAtt\_depth. This validates the effectiveness of PDP, and also shows that simply introducing the `outside token' and patch attention module without the supervision of patch distribution loss will hurt the performance. We also tried additionally predicting a 2D gaze direction from the head feature $f_h$, but did not observe any improvement on GazeFollow. 

\subsubsection{Performance Regarding Annotation Variance}

We evaluated our model's performance on images with larger annotation variance in the GazeFollow test set. To compute the annotation variance score for each image, we calculated the mean of the distances between each annotated coordinate and the average coordinate of all annotations. A larger distance indicates more disagreement between annotators. The statistics of the variance scores can be seen in Figure \ref{fig:gazefollow_info}. We divided the dataset into 10 equal parts according to the quantiles of the variance score, each containing about 450 images. The model is evaluated in each part, both with and without depth maps as input, and compared with the corresponding version of the VideoAtt model. As we expect to examine the distribution of the heatmap predictions, we focus on the AUC metric.

\begin{figure}[t]
\centering
 \includegraphics[width=\linewidth, height=1.1in]{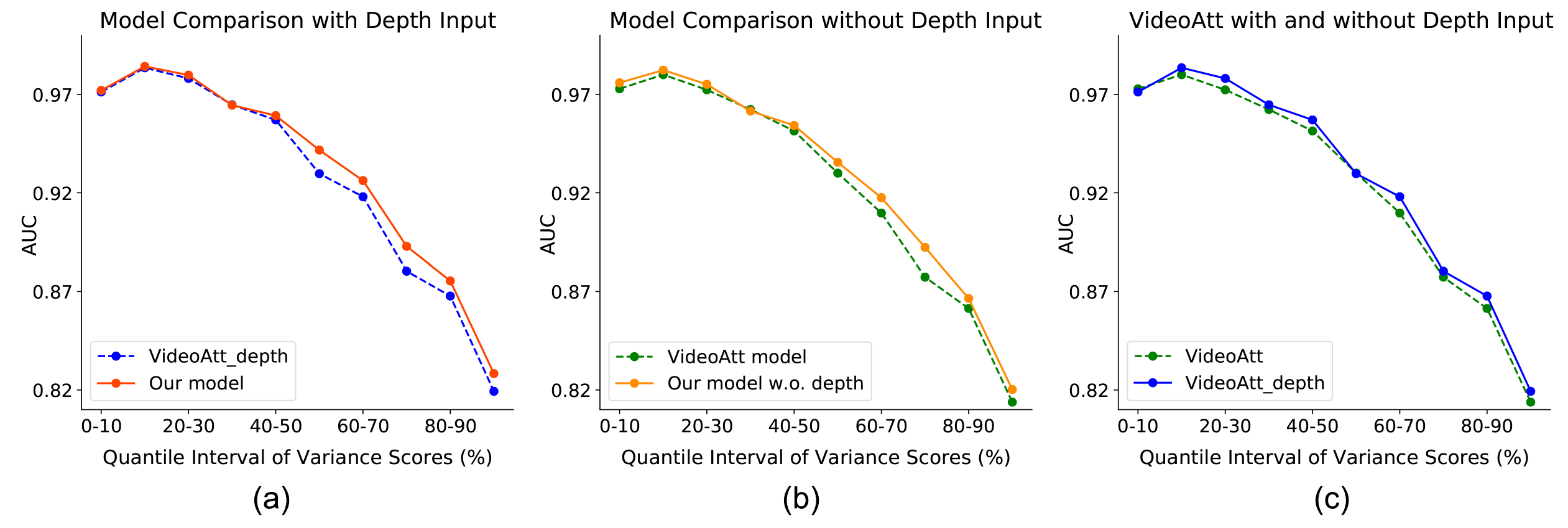}

\caption{Comparison of AUC in each quantile interval in the GazeFollow test set. Our model shows obviously higher AUC in intervals with larger annotation variance, which cannot be observed in (c) when adding depth input only.}
\label{fig:uncertain_compare}
\end{figure}

Figure \ref{fig:uncertain_compare} shows our comparison results. It can be observed that our model shows an obviously higher AUC on images with variance scores above the 50\% quantile, while the performances of our model and the VideoAtt model are highly close for images with variance scores below the 50\% quantiles. To rule out the potential effect of other factors, we also computed the differences in performance between the VideoAtt\_depth and VideoAtt model. Results show that the performance gain by incorporating depth maps as input lies evenly in the upper and lower quantiles, which in turn substantiates the claim that the performance gain from our method comes from the better heatmap predictions for images with larger variance in annotations by considering the uncertainty in gaze following annotations.

\begin{figure}[t]
\setlength\tabcolsep{2pt}%%
\renewcommand{\arraystretch}{1} 
\centering
\begin{tabular}{ccc}
 \textbf{Input} &
 \textbf{Videoatt\_depth} &
 \textbf{Our model} \\
 \includegraphics[width=0.32\linewidth, height=0.75in]{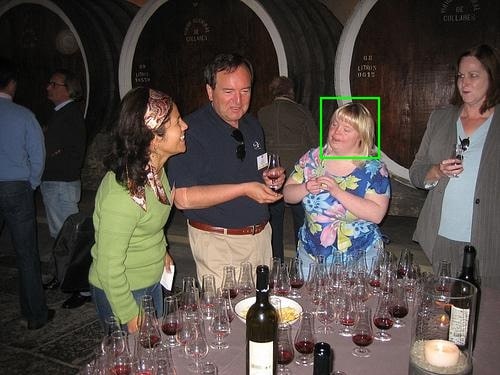} &
 \includegraphics[width=0.32\linewidth, height=0.75in]{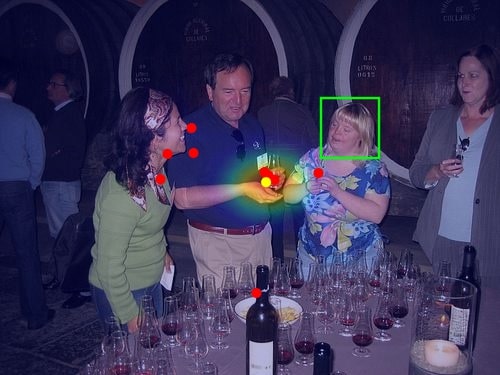} &
 \includegraphics[width=0.32\linewidth, height=0.75in]{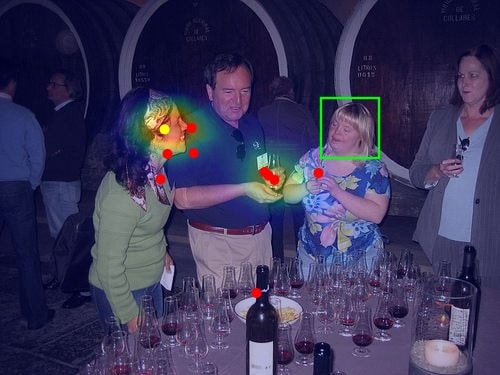} \\
 \includegraphics[width=0.32\linewidth, height=0.75in]{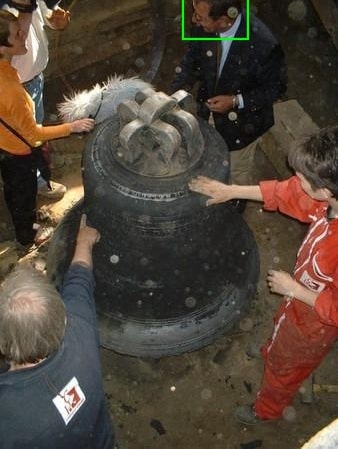} &
 \includegraphics[width=0.32\linewidth, height=0.75in]{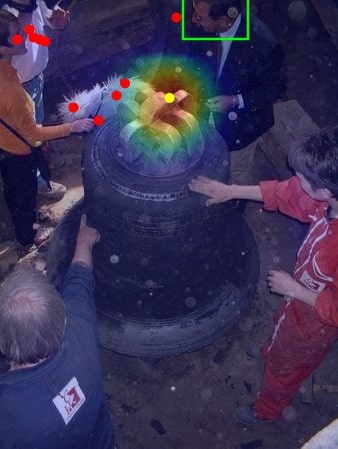} &
 \includegraphics[width=0.32\linewidth, height=0.75in]{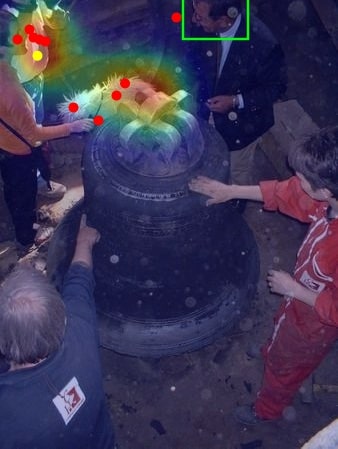} \\
 \includegraphics[width=0.32\linewidth, height=0.75in]{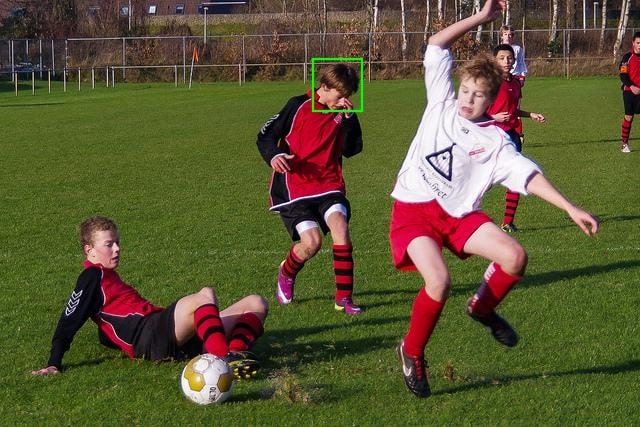} &
 \includegraphics[width=0.32\linewidth, height=0.75in]{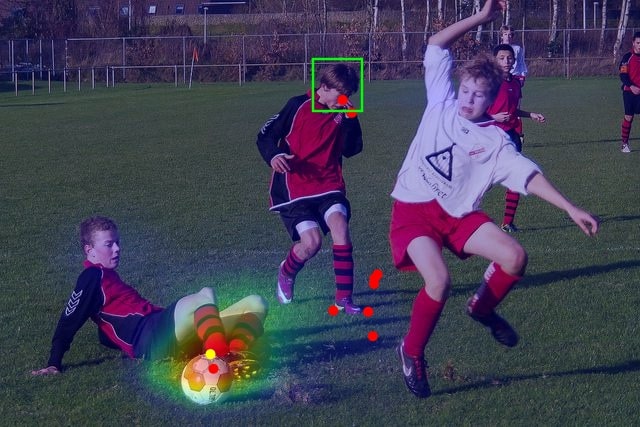} &
 \includegraphics[width=0.32\linewidth, height=0.75in]{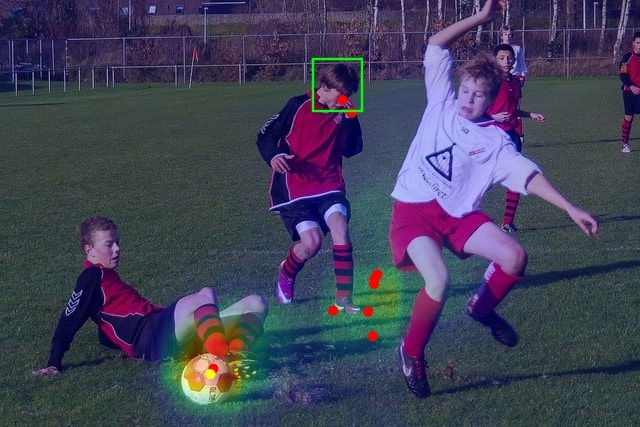} 
\end{tabular}
\caption{Predicted heatmaps from our model and VideoAtt\_depth on images with larger ``variance scores". The predicted (yellow) and ground truth (red) target points are also plotted. Our predicted heatmaps are more aligned with the group-level annotations.} 
\label{fig:gazefollow_vis}
\end{figure}

Figure \ref{fig:gazefollow_vis} visualizes the predicted heatmaps of our model and the VideoAtt\_depth Model on some example images with larger variance score. In contrast to the VideoAtt\_depth model which only predicts a unimodal Gaussian, our model predicts multi-modal heatmap predictions which are better aligned with the group-level human annotations.

\subsubsection{Comparison w/ Original In/Out Prediction Task}
As mentioned earlier, when training on VideoAttentionTarget, the previous models \cite{chong2020detecting,fang2021dual, Bao_2022_CVPR} first pretrain the model on GazeFollow dataset, using MSE loss for the heatmap prediction task, and BCE loss for the in/out prediction task. However, as claimed in the official code of the VideoAtt model, in order to get SOTA performance on GazeFollow, the BCE loss was not involved in training. In our experiments, we found that when the model is trained with two losses together, there is a large drop in performance for the heatmap prediction task, as in Table \ref{tab:inout_check}. This seemingly weird result may stem from the separate handling of the two subtasks, causing introducing the in/out prediction task hurting the performance in the target prediction task. 

\begin{table}[htbp]
    \centering
    \caption{Effect of the In/out Prediction Task on Heatmap Prediction Task for VideoAtt model on GazeFollow dataset}
    \begin{tabular}{l c c c}
        \toprule
        \multirow{2}{*}{Method} & \multirow{2}{*}{ AUC $\uparrow$} & \multicolumn{2}{c}{ Dist. $\downarrow$}  \\
                        &                     & Avg. & Min. \\
        \toprule
        VideoAtt \cite{chong2020detecting} & 0.921 & 0.137 & 0.077 \\
        VideoAtt w. in/out & 0.921 & 0.147 & 0.083 \\
        VideoAtt\_depth & 0.927 & 0.131 & 0.071 \\
        VideoAtt\_depth w. in/out & 0.927 & 0.145 & 0.083 \\
        \toprule
        Ours w.o. dep. & 0.928 & 0.131 & 0.072 \\
        Ours & \textbf{0.934} & \textbf{0.123} & \textbf{0.065} \\
        \bottomrule
    \end{tabular}
    \label{tab:inout_check}
\end{table}

In contrast, our PDP method integrates the two subtasks without loss in performance, which also enables a much more efficient way of pretraining. Our model only needs to be trained once, instead of training two versions of the model to get the best performance on the GazeFollow dataset, and VideoAttentionTarget dataset respectively.

\section{Conclusions} \label{conclu}

In this paper, we propose the PDP method in gaze following. By using the extracted feature vectors as inside tokens and adding an outside token, a patch-level gaze distribution is predicted. The PDP method can serve as a regularization method to the MSE loss for heatmap regression. Experiments show the superior performance of our method over the baseline models with an obviously better performance on images with larger annotation variance. Furthermore, the PDP task bridges the gap between the target prediction and in/out prediction tasks by showing a significantly higher AP, and provides a much simpler way of training gaze following models for any in-the-wild images/videos.

\paragraph{Acknowledgements.} This project was partially supported by US National Science Foundation Awards IIS-1763981, IIS-2123920, DUE-2055406, the Partner University Fund, the SUNY2020 Infrastructure Transportation Security Center, and a gift from Adobe.

{\small
\bibliographystyle{ieee_fullname}
\bibliography{main}

@inproceedings{Rebello-etal-PERC19,
author = {Rebello, Sanjay and Hoai, Minh and Wang, Yang and Zu, Tianlong and Hutson, John and Loschky, Lester},
year = {2018},
title = {Machine Learning Predicts Responses to Conceptual Questions Using Eye Movements},
booktitle = {Proceedings of the Physics Education Research Conference (PERC)},
}

@article{emery2000eyes,
  title={The eyes have it: the neuroethology, function and evolution of social gaze},
  author={Emery, Nathan J},
  journal={Neuroscience \& biobehavioral reviews},
  volume={24},
  number={6},
  pages={581--604},
  year={2000},
  publisher={Elsevier}
}

@article{abele1986functions,
  title={Functions of gaze in social interaction: Communication and monitoring},
  author={Abele, Andrea},
  journal={Journal of Nonverbal Behavior},
  volume={10},
  number={2},
  pages={83--101},
  year={1986},
  publisher={Springer}
}

@article{pfeiffer2012eyes,
  title={Eyes on the mind: investigating the influence of gaze dynamics on the perception of others in real-time social interaction},
  author={Pfeiffer, Ulrich and Schilbach, Leonhard and Timmermans, Bert and Jording, Mathis and Bente, Gary and Vogeley, Kai},
  journal={Frontiers in psychology},
  volume={3},
  pages={537},
  year={2012},
  publisher={Frontiers}
}

@article{charman1997infants,
  title={Infants with autism: an investigation of empathy, pretend play, joint attention, and imitation.},
  author={Charman, Tony and Swettenham, John and Baron-Cohen, Simon and Cox, Antony and Baird, Gillian and Drew, Auriol},
  journal={Developmental psychology},
  volume={33},
  number={5},
  pages={781},
  year={1997},
  publisher={American Psychological Association}
}

@article{morimoto2005eye,
  title={Eye gaze tracking techniques for interactive applications},
  author={Morimoto, Carlos H and Mimica, Marcio RM},
  journal={Computer vision and image understanding},
  volume={98},
  number={1},
  pages={4--24},
  year={2005},
  publisher={Elsevier}
}

@incollection{majaranta2014eye,
  title={Eye tracking and eye-based human--computer interaction},
  author={Majaranta, P{\"a}ivi and Bulling, Andreas},
  booktitle={Advances in physiological computing},
  pages={39--65},
  year={2014},
  publisher={Springer}
}

@article{jaswal2019being,
  title={Being versus appearing socially uninterested: Challenging assumptions about social motivation in autism},
  author={Jaswal, Vikram K and Akhtar, Nameera},
  journal={Behavioral and Brain Sciences},
  volume={42},
  year={2019},
  publisher={Cambridge University Press}
}

@book{baron1997mindblindness,
  title={Mindblindness: An essay on autism and theory of mind},
  author={Baron-Cohen, Simon},
  year={1997},
  publisher={MIT press}
}

@inproceedings{NIPS2015_ec895663,
 author = {Recasens, Adria and Khosla, Aditya and Vondrick, Carl and Torralba, Antonio},
 booktitle = {Advances in Neural Information Processing Systems},
 editor = {C. Cortes and N. Lawrence and D. Lee and M. Sugiyama and R. Garnett},
 pages = {},
 publisher = {Curran Associates, Inc.},
 title = {Where are they looking?},
 url = {https://proceedings.neurips.cc/paper/2015/file/ec8956637a99787bd197eacd77acce5e-Paper.pdf},
 volume = {28},
 year = {2015}
}

@inproceedings{dias2020gaze,
  title={Gaze estimation for assisted living environments},
  author={Dias, Philipe Ambrozio and Malafronte, Damiano and Medeiros, Henry and Odone, Francesca},
  booktitle={Proceedings of the IEEE/CVF Winter Conference on Applications of Computer Vision},
  pages={290--299},
  year={2020}
}

@inproceedings{recasens2017following,
  title={Following gaze in video},
  author={Recasens, Adria and Vondrick, Carl and Khosla, Aditya and Torralba, Antonio},
  booktitle={Proceedings of the IEEE International Conference on Computer Vision},
  pages={1435--1443},
  year={2017}
}

@inproceedings{chong2018connecting,
  title={Connecting gaze, scene, and attention: Generalized attention estimation via joint modeling of gaze and scene saliency},
  author={Chong, Eunji and Ruiz, Nataniel and Wang, Yongxin and Zhang, Yun and Rozga, Agata and Rehg, James M},
  booktitle={Proceedings of the European conference on computer vision (ECCV)},
  pages={383--398},
  year={2018}
}

@inproceedings{lian2018believe,
  title={Believe It or Not, We Know What You Are Looking At!},
  author={Lian, Dongze and Yu, Zehao and Gao, Shenghua},
  booktitle={Asian Conference on Computer Vision},
  pages={35--50},
  year={2018},
  organization={Springer}
}

@inproceedings{chong2020detecting,
  title={Detecting attended visual targets in video},
  author={Chong, Eunji and Wang, Yongxin and Ruiz, Nataniel and Rehg, James M},
  booktitle={Proceedings of the IEEE/CVF Conference on Computer Vision and Pattern Recognition},
  pages={5396--5406},
  year={2020}
}

@inproceedings{fang2021dual,
  title={Dual Attention Guided Gaze Target Detection in the Wild},
  author={Fang, Yi and Tang, Jiapeng and Shen, Wang and Shen, Wei and Gu, Xiao and Song, Li and Zhai, Guangtao},
  booktitle={Proceedings of the IEEE/CVF Conference on Computer Vision and Pattern Recognition},
  pages={11390--11399},
  year={2021}
}

@inproceedings{jin2021multi,
  title={Multi-Person Gaze-Following with Numerical Coordinate Regression},
  author={Jin, Tianlei and Lin, Zheyuan and Zhu, Shiqiang and Wang, Wen and Hu, Shunda},
  booktitle={2021 16th IEEE International Conference on Automatic Face and Gesture Recognition (FG 2021)},
  pages={01--08},
  year={2021},
  organization={IEEE}
}

@inproceedings{nakazawa2012point,
  title={Point of gaze estimation through corneal surface reflection in an active illumination environment},
  author={Nakazawa, Atsushi and Nitschke, Christian},
  booktitle={European Conference on Computer Vision},
  pages={159--172},
  year={2012},
  organization={Springer}
}

@article{shih2004novel,
  title={A novel approach to 3-D gaze tracking using stereo cameras},
  author={Shih, Sheng-Wen and Liu, Jin},
  journal={IEEE Transactions on Systems, Man, and Cybernetics, Part B (Cybernetics)},
  volume={34},
  number={1},
  pages={234--245},
  year={2004},
  publisher={IEEE}
}

@article{guestrin2006general,
  title={General theory of remote gaze estimation using the pupil center and corneal reflections},
  author={Guestrin, Elias Daniel and Eizenman, Moshe},
  journal={IEEE Transactions on biomedical engineering},
  volume={53},
  number={6},
  pages={1124--1133},
  year={2006},
  publisher={IEEE}
}

@inproceedings{zhu2005eye,
  title={Eye gaze tracking under natural head movements},
  author={Zhu, Zhiwei and Ji, Qiang},
  booktitle={2005 IEEE Computer Society Conference on Computer Vision and Pattern Recognition (CVPR'05)},
  volume={1},
  pages={918--923},
  year={2005},
  organization={IEEE}
}

@inproceedings{zhang2015appearance,
  title={Appearance-based gaze estimation in the wild},
  author={Zhang, Xucong and Sugano, Yusuke and Fritz, Mario and Bulling, Andreas},
  booktitle={Proceedings of the IEEE conference on computer vision and pattern recognition},
  pages={4511--4520},
  year={2015}
}

@inproceedings{fischer2018rt,
  title={Rt-gene: Real-time eye gaze estimation in natural environments},
  author={Fischer, Tobias and Chang, Hyung Jin and Demiris, Yiannis},
  booktitle={Proceedings of the European Conference on Computer Vision (ECCV)},
  pages={334--352},
  year={2018}
}

@inproceedings{cheng2018appearance,
  title={Appearance-based gaze estimation via evaluation-guided asymmetric regression},
  author={Cheng, Yihua and Lu, Feng and Zhang, Xucong},
  booktitle={Proceedings of the European Conference on Computer Vision (ECCV)},
  pages={100--115},
  year={2018}
}

@inproceedings{krafka2016eye,
  title={Eye tracking for everyone},
  author={Krafka, Kyle and Khosla, Aditya and Kellnhofer, Petr and Kannan, Harini and Bhandarkar, Suchendra and Matusik, Wojciech and Torralba, Antonio},
  booktitle={Proceedings of the IEEE conference on computer vision and pattern recognition},
  pages={2176--2184},
  year={2016}
}

@inproceedings{park2019few,
  title={Few-shot adaptive gaze estimation},
  author={Park, Seonwook and Mello, Shalini De and Molchanov, Pavlo and Iqbal, Umar and Hilliges, Otmar and Kautz, Jan},
  booktitle={Proceedings of the IEEE/CVF International Conference on Computer Vision},
  pages={9368--9377},
  year={2019}
}

@inproceedings{kellnhofer2019gaze360,
  title={Gaze360: Physically unconstrained gaze estimation in the wild},
  author={Kellnhofer, Petr and Recasens, Adria and Stent, Simon and Matusik, Wojciech and Torralba, Antonio},
  booktitle={Proceedings of the IEEE/CVF International Conference on Computer Vision},
  pages={6912--6921},
  year={2019}
}

@article{koenker2001quantile,
  title={Quantile regression},
  author={Koenker, Roger and Hallock, Kevin F},
  journal={Journal of economic perspectives},
  volume={15},
  number={4},
  pages={143--156},
  year={2001}
}

@article{Ranftl2020,
	author    = {Ren\'{e} Ranftl and Katrin Lasinger and David Hafner and Konrad Schindler and Vladlen Koltun},
	title     = {Towards Robust Monocular Depth Estimation: Mixing Datasets for Zero-shot Cross-dataset Transfer},
	journal   = {IEEE Transactions on Pattern Analysis and Machine Intelligence (TPAMI)},
	year      = {2020},
}

@article{ramachandran2019stand,
  title={Stand-alone self-attention in vision models},
  author={Ramachandran, Prajit and Parmar, Niki and Vaswani, Ashish and Bello, Irwan and Levskaya, Anselm and Shlens, Jonathon},
  journal={arXiv preprint arXiv:1906.05909},
  year={2019}
}

@inproceedings{he2016deep,
  title={Deep residual learning for image recognition},
  author={He, Kaiming and Zhang, Xiangyu and Ren, Shaoqing and Sun, Jian},
  booktitle={Proceedings of the IEEE conference on computer vision and pattern recognition},
  pages={770--778},
  year={2016}
}

@inproceedings{funes2014eyediap,
  title={Eyediap: A database for the development and evaluation of gaze estimation algorithms from rgb and rgb-d cameras},
  author={Funes Mora, Kenneth Alberto and Monay, Florent and Odobez, Jean-Marc},
  booktitle={Proceedings of the Symposium on Eye Tracking Research and Applications},
  pages={255--258},
  year={2014}
}

@article{zhou2014learning,
  title={Learning deep features for scene recognition using places database},
  author={Zhou, Bolei and Lapedriza, Agata and Xiao, Jianxiong and Torralba, Antonio and Oliva, Aude},
  year={2014},
  publisher={Neural Information Processing Systems Foundation}
}

@article{kingma2014adam,
  title={Adam: A method for stochastic optimization},
  author={Kingma, Diederik P and Ba, Jimmy},
  journal={arXiv preprint arXiv:1412.6980},
  year={2014}
}

@inproceedings{zhang2018coarse,
  title={From Coarse Attention to Fine-Grained Gaze: A Two-stage 3D Fully Convolutional Network for Predicting Eye Gaze in First Person Video.},
  author={Zhang, Zehua and Crandall, David J and Yu, Chen and Bambach, Sven},
  booktitle={BMVC},
  pages={295},
  year={2018}
}

@InProceedings{Tu_2022_CVPR,
    author    = {Tu, Danyang and Min, Xiongkuo and Duan, Huiyu and Guo, Guodong and Zhai, Guangtao and Shen, Wei},
    title     = {End-to-End Human-Gaze-Target Detection With Transformers},
    booktitle = {Proceedings of the IEEE/CVF Conference on Computer Vision and Pattern Recognition (CVPR)},
    month     = {June},
    year      = {2022},
    pages     = {2202-2210}
}

@InProceedings{Bao_2022_CVPR,
    author    = {Bao, Jun and Liu, Buyu and Yu, Jun},
    title     = {ESCNet: Gaze Target Detection With the Understanding of 3D Scenes},
    booktitle = {Proceedings of the IEEE/CVF Conference on Computer Vision and Pattern Recognition (CVPR)},
    month     = {June},
    year      = {2022},
    pages     = {14126-14135}
}

@inproceedings{masse2019extended,
  title={Extended gaze following: Detecting objects in videos beyond the camera field of view},
  author={Mass{\'e}, Benoit and Lathuili{\`e}re, St{\'e}phane and Mesejo, Pablo and Horaud, Radu},
  booktitle={2019 14th IEEE International Conference on Automatic Face \& Gesture Recognition (FG 2019)},
  pages={1--8},
  year={2019},
  organization={IEEE}
}

@inproceedings{tomas2021goo,
  title={Goo: A dataset for gaze object prediction in retail environments},
  author={Tomas, Henri and Reyes, Marcus and Dionido, Raimarc and Ty, Mark and Mirando, Jonric and Casimiro, Joel and Atienza, Rowel and Guinto, Richard},
  booktitle={Proceedings of the IEEE/CVF Conference on Computer Vision and Pattern Recognition},
  pages={3125--3133},
  year={2021}
}

@inproceedings{wang2022gatector,
  title={GaTector: A Unified Framework for Gaze Object Prediction},
  author={Wang, Binglu and Hu, Tao and Li, Baoshan and Chen, Xiaojuan and Zhang, Zhijie},
  booktitle={Proceedings of the IEEE/CVF Conference on Computer Vision and Pattern Recognition},
  pages={19588--19597},
  year={2022}
}

@article{bhattacharyya1943measure,
  title={On a measure of divergence between two statistical populations defined by their probability distributions},
  author={Bhattacharyya, Anil},
  journal={Bull. Calcutta Math. Soc.},
  volume={35},
  pages={99--109},
  year={1943}
}

@article{lin1991divergence,
  title={Divergence measures based on the Shannon entropy},
  author={Lin, Jianhua},
  journal={IEEE Transactions on Information theory},
  volume={37},
  number={1},
  pages={145--151},
  year={1991},
  publisher={IEEE}
}
}

%%%%%%%%%% Merge with supplemental materials %%%%%%%%%%

\onecolumn
\begin{center}
\textbf{\large Patch-level Gaze Distribution Prediction for Gaze Following: Supplementary Material}
\end{center}

\setcounter{equation}{0}
\setcounter{figure}{0}
\setcounter{table}{0}
\setcounter{section}{0}
\makeatletter
\renewcommand{\theequation}{S\arabic{equation}}
\renewcommand{\thefigure}{S\arabic{figure}}
\renewcommand{\thetable}{S\arabic{table}}
\renewcommand{\thesection}{S{\arabic{section}}}
\renewcommand{\theHsection}{S\arabic{section}}
\renewcommand\thesubsection{\thesection.\arabic{subsection}}
\vspace{5mm}

In this supplementary material, we provide additional information for further understanding of our method:
\begin{itemize}
   \item Section \hyperref[sec:imp]{S1} provides the implementation details. 
  \item 
  Section \ref{sec:feat_extractor} explains the detailed structure of the feature extraction module. 
 \item Section \ref{sec:patchgen_test} shows the results of different alternatives to the patch distribution creation method. 
 \item Section \ref{sec:pred_consist} analyzed the level of consistency between the predicted heatmaps and patch distributions. 
 \item Section \ref{sec:depth_comparison} shows the comparison between our model and the VideoAtt model both with and without depth input. 
 \item Section \ref{sec:failure_cases} shows some example failure cases of our model.
\end{itemize}

\section{Implementation Details} \label{sec:imp}
In all our experiments, all input images are resized to $224 \times 224$. Both the scene backbone and head backbone are ResNet-50 \cite{he2016deep} followed by an additional residual layer and average pooling layer for dimensionality reduction. The output feature dimensions from both backbones are $1024 \times 7 \times 7$. Same with the VideoAtt model \cite{chong2020detecting}, the head backbone is initialized with weights pretrained on the Eyediap dataset \cite{funes2014eyediap}, and the scene backbone is initialized with pretrained weights on the Places dataset \cite{zhou2014learning}. The encoder has two convolutional layers with kernel sizes of $1 \times 1$, which reduce the channels from 2048 to 512. Therefore, the number of patch tokens is $7 \times 7 + 1 = 50$. We set $\sigma = 3$  for generating the ground truth heatmap following the default setting of previous models \cite{lian2018believe, chong2020detecting}. 

As the procedure used in VideoAtt \cite{chong2020detecting} and DualAtt \cite{fang2021dual}, the model is first trained on the GazeFollow dataset until convergence, and then finetuned on the VideoAttentionTarget dataset. Adam \cite{kingma2014adam} was used to optimize the model with a learning rate of 2.5e-4, which is decreased with a decay factor of 0.2 at the 25th, 31st, and 40th epochs on the GazeFollow dataset. For finetuning on the VideoAttentionTarget dataset, we used a 5-frame sequence as one sample. The weights until the patch attention module are frozen, and the rest of the network is trained with a learning rate of 1e-4, with a decay factor of 0.5 at the 3rd and 6th epochs. The batch sizes for training on GazeFollow and VideoAttentionTarget are 80 and 16 respectively.

\section{Structure of the Feature Extraction Module} \label{sec:feat_extractor}

\begin{figure}[t!]
    \centering
    \includegraphics[width=0.8\linewidth, height=2.8in]{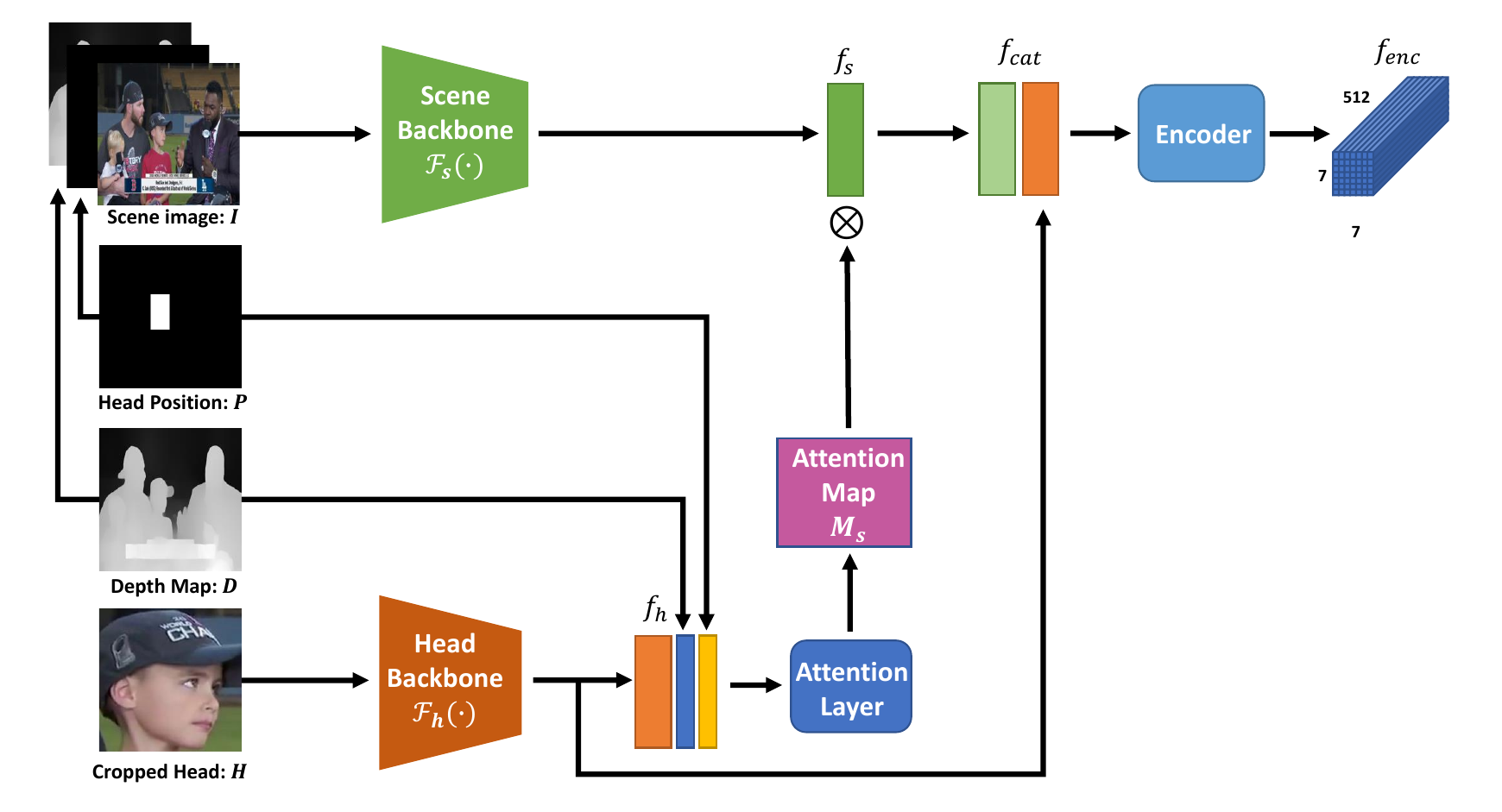}
    \caption{Structure of the feature extraction module. $\otimes$ indicates element-wise multiplication. Please see text for details.}
    \label{fig:detailed_model}
\end{figure}

The structure of the feature extraction module is shown in Figure \ref{fig:detailed_model}. We leverage the feature extraction component of the VideoAtt model \cite{chong2020detecting} for feature extraction, with some small modifications. The feature extraction module consists of two branches: a scene branch and a head branch. In the scene branch, the scene backbone $\mathcal{F}_s(\cdot)$, takes the scene image $I \in R^{3 \times H_0 \times W_0}$, the binary head position mask $P  \in \{0, 1\}^{H_0 \times W_0}$, and a normalized depth map $D \in [0,1]^{H_0 \times W_0}$ as input, and output the scene feature $f_s \in R^{C \times H \times W}$. We leveraged an additional depth map as input according to the insight from the DualAtt model \cite{fang2021dual} to incorporate scene depth information, which is computed with an off-the-shelf monocular depth estimation model \cite{Ranftl2020}. The head backbone $\mathcal{F}_h(\cdot)$ takes the head crop of the person $H \in R^{3 \times H_0 \times W_0}$  as input, and output the head feature $f_h \in R^{C \times H \times W}$. The average pooled head feature is concatenated with the downsampled head mask and the depth map and fed into an attention layer to generate a spatial attention map $M_s \in R^{H \times W}$, which is multiplied with the scene feature and concatenated with the face feature:
\begin{equation}
    f_{cat} = [M_s \otimes f_s, f_h],
\end{equation}
where $[\cdot, \cdot]$ denotes concatenation operation, and $\otimes$ denotes element-wise multiplication on each channel in $f_s$. The Attention layer is a fully connected layer mapping from the dimension of the concatenated feature to the size of $H \times W$. The VideoAtt model uses two separate encoders for the heatmap prediction and in/out prediction branch after $f_{cat}$. In contrast, we used a single encoder with two convolutional layers to extract shared feature encoding $f_{enc} \in R^{C \times H \times W}$ , as we expect the PDP task to benefit the learning of the shared feature by merging the two subtasks. 

\section{Ablations of the Ground Truth Patch Distribution Creation Method} \label{sec:patchgen_test}

In Table \ref{tab:patch_ablation} we provide the results of ablations of the ground truth patch distribution creation method. First, we tested choosing different numbers of patches for the patch-level gaze distribution. As the number of patches corresponds to the number of tokens in $f_{enc}$ after the feature extraction module, we only tested in the scale of 2 ($4 \times 4$ and $14 \times 14$) for ease of implementation. Setting the patch number to $4 \times 4$ has a large drop in performance on GazeFollow, possibly because the overly coarse scale feature encodings make the target estimation in finer grain difficult. When using a patch number of $14 \times 14$, despite lower drop in performance on GazeFollow, the model performance on VideoAttentionTarget drops obviously. We infer that this is because when training on VideoAttentionTarget (which has lots of outside cases), the much larger number of inside tokens (4 times of $7 \times 7$) make the patch distribution prediction difficult when combined with the outside token. 

\begin{table*}[htbp]
\centering
\begin{tabular}{lcccccc}
\toprule
\multirow{3}{*}{Method} & \multicolumn{3}{c}{GazeFollow}                   & \multicolumn{3}{c}{VideoAttentionTarget}                                                       \\ \cmidrule(lr){2-4} \cmidrule(lr){5-7}  
                        & \multirow{2}{*}{AUC $\uparrow$} & \multicolumn{2}{c}{Dist. $\downarrow$} & \multicolumn{2}{c}{\textit{In frame}}              & \multicolumn{1}{c}{\textit{Out of frame}} \\
                        &                      & Avg.        & Min.        & \multicolumn{1}{c}{AUC $\uparrow$} & \multicolumn{1}{c}{Dist $\downarrow$} & \multicolumn{1}{c}{AP $\uparrow$} \\ \toprule
Patch: $4 \times 4$    & 0.921                                    & 0.129                                          & 0.071   & 0.911        & 0.110                 & 0.897                                                        \\

Patch: $14 \times 14$   &  0.930                                   & 0.127                                          & 0.067  & 0.908     & 0.113                 & 0.892                                                           \\
MaxPool $\rightarrow$ AvgPool          &  0.931                                    &  0.124                                              & 0.066    & 0.913         & 0.112        & 0.903    \\
One-hot       & 0.925                                    & 0.128 & 0.071                                           & 0.903             & 0.110                     & 0.884                                                              \\
\toprule
Ours         & \textbf{0.934} &\textbf{0.123}                                   & \textbf{0.065}                                       & \textbf{0.917}         & \textbf{0.109}                 & \textbf{0.908}          \\
\bottomrule
\end{tabular}
\caption{Ablations of Patch Distribution Creation Method}
\label{tab:patch_ablation}
\end{table*}

\begin{figure}[htbp]
    \centering
    \includegraphics[width=\linewidth]{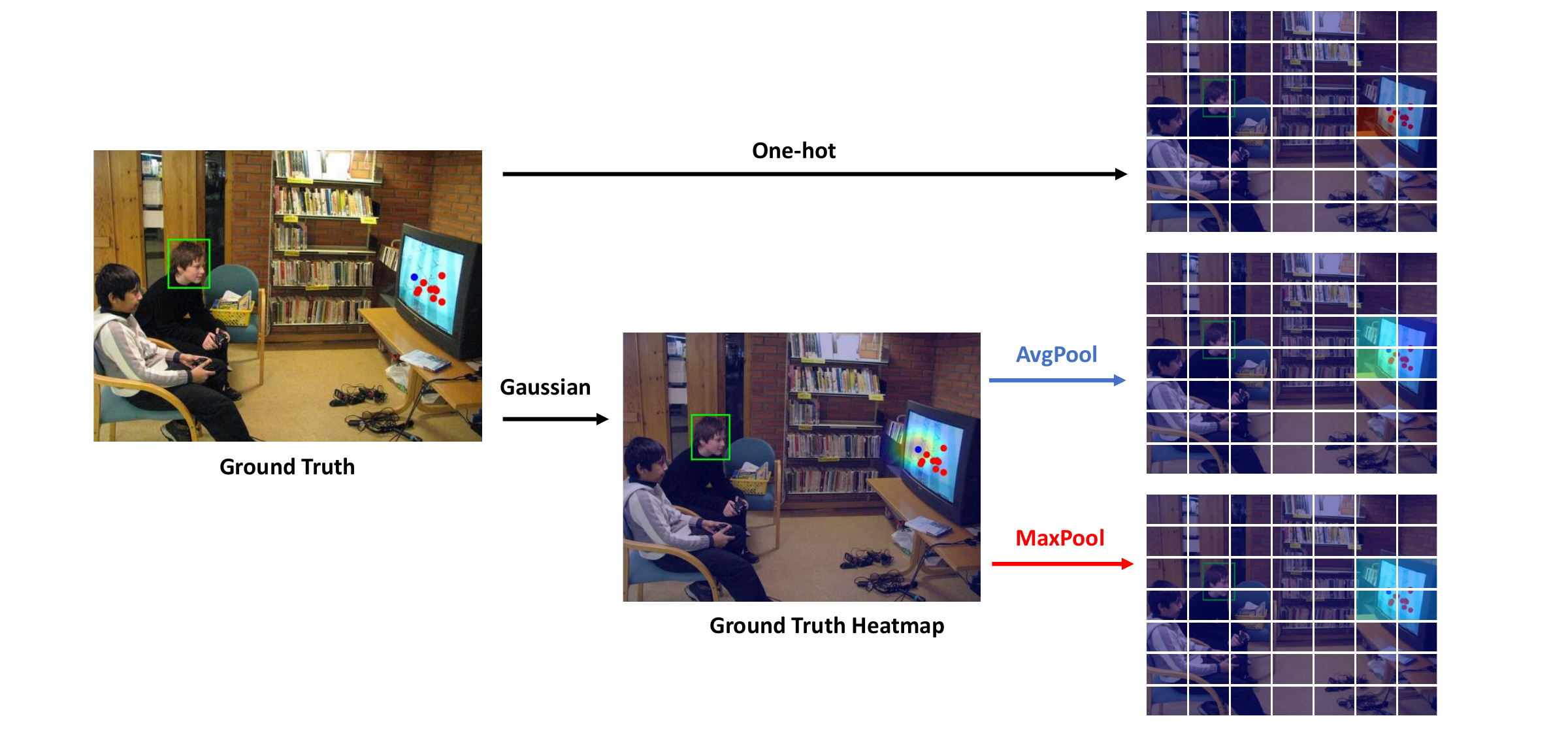}
    \caption{Visualizations of patch distributions generated in different ways from a sampled ground truth annotation (blue) on an example image in the GazeFollow test set. Annotations from other annotators are visualized in red. In the case of an annotation point being close to the patch boundary, the one-hot design simply regards the neighboring patches as unattended, while our max pooling method can generate much better distribution with high response in different patches, encouraging the model to predict multi-modal heatmaps in inference.}
    \label{fig:patch_dist_abl}
\end{figure}

We also trained the model using average pooling instead of max pooling to get the ground truth patch distribution. The model still shows good performance but with a slight drop in all metrics. Finally, replacing our PDP task with the one-hot patch classification as in \cite{zhang2018coarse} caused a significant drop in performance, showing little improvement compared to the VideoAtt\_depth model.
In Figure \ref{fig:patch_dist_abl}, we also visualized the distributions generated by different alternatives from one sampled annotation (blue) on an example image in the GazeFollow test set. We assume in training, only the sampled annotation is available.  It can be seen that the one-hot generation method assigns a hard label of 1 for the specific patch in which the annotated point is located, which is unimodal and cannot cover the adjacent patches if the target is located close to the patch boundaries. This limits the model's capability to predict multi-modal heatmaps.  When discretizing the ground truth heatmap with average pooling, a smoother patch distribution can be obtained. However, the patch where the point is located still has a much higher response than the right neighboring patches, despite the right patches being very close to the point. Our method of max pooling generates the distribution that  aligns well with the group-level human annotations by creating higher confidences in multiple patches, showing the best potential to make the model generate multi-modal outputs. 

\section{Consistency between the  Heatmap and Patch Distribution Predictions} \label{sec:pred_consist}
As mentioned in the main paper, we used PDP as a regularization method for heatmap prediction, and the predicted heatmap is used as the final output for the target prediction task if the target is located inside the image. However, after visualizing the outputs of our model as in Figure 6 in the main paper, we found high consistency between the predicted heatmaps and the patch distribution. To further investigate the level of consistency between the predictions, we created patch distributions from the predicted heatmaps using our method for creating the ground truth patch distribution from the ground truth heatmap. We call this created patch distribution as ``Patch Distribution from Predicted Heatmap" (PDPH). We computed the similarities and distances between the patch distribution predicted from the patch prediction head and PDPH from the heatmap prediction head on the test set of GazeFollow and VideoAttentionTarget. We selected the Bhattacharyya coefficient \cite{bhattacharyya1943measure} and the Jensen-Shannon (JS) divergence \cite{lin1991divergence} as the evaluation metrics due to their suitability for computing similarity or distances between distributions. Note that we used JS divergence here instead of KL divergence due to its symmetrical property as we do not have `ground truth' patch distribution here.

As shown in Table \ref{tab:corres_analysis}, the Bhattacharyya coefficients show very high value on both datasets while the values of JS divergence are very low. This high consistency between the outputs further demonstrates that our model design and the ground truth patch distribution creation method can regularize the heatmap prediction by acting on the common feature embedding before the prediction heads.

\begin{table}[htbp]
    \centering
    \begin{tabular}{l c c c}
     \toprule
     Dataset & Bhat. Coef. $\uparrow$ & JS Div. $\downarrow$ \\
     \toprule
     GazeFollow & 0.976 & 0.142 \\
     VideoAttentionTarget & 0.971 & 0.158 \\
     \bottomrule
    \end{tabular}
    \centering
    \caption{Analysis of Consistencies between Patch and Heatmap Predictions.}
\label{tab:corres_analysis}
\end{table}

\section{Model Performance with and without Depth} \label{sec:depth_comparison}
\begin{figure*}[htbp]
\setlength\tabcolsep{1pt}%%
\centering
\begin{tabular}{cccc}
\makecell{\textbf{VideoAtt}} &
 \makecell{\textbf{Ours} \\ \textbf{w.o. depth}} &
 \makecell{\textbf{VideoAtt\_depth}} &
 \makecell{\textbf{Ours}}\\
 \includegraphics[width=0.24\linewidth, height=0.24\linewidth]{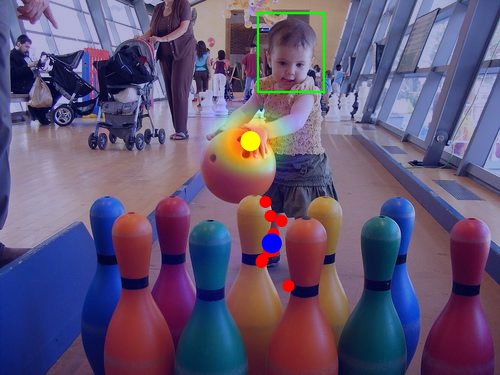} &
 \includegraphics[width=0.24\linewidth, height=0.24\linewidth]{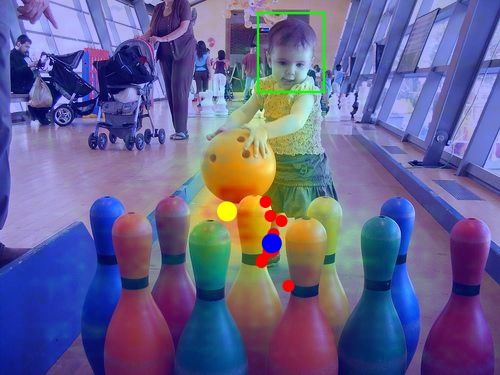} &
 \includegraphics[width=0.24\linewidth, height=0.24\linewidth]{images/supp/4models/gf_base_1.jpg}&
  \includegraphics[width=0.24\linewidth, height=0.24\linewidth]{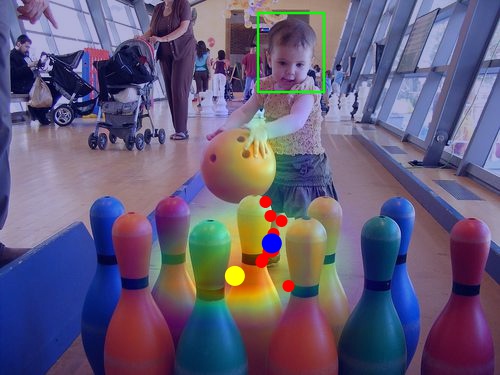}\\
 \includegraphics[width=0.24\linewidth, height=0.24\linewidth]{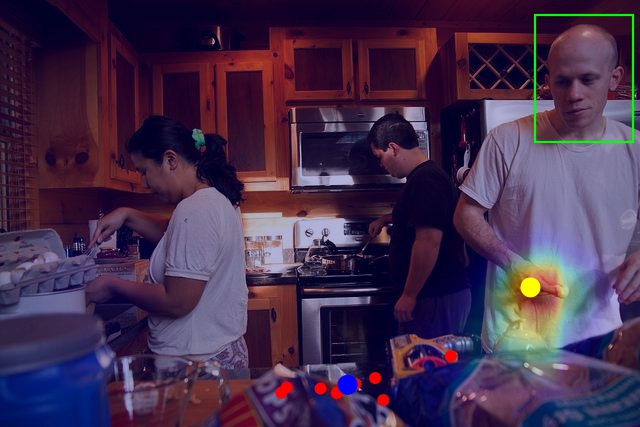} &
 \includegraphics[width=0.24\linewidth, height=0.24\linewidth]{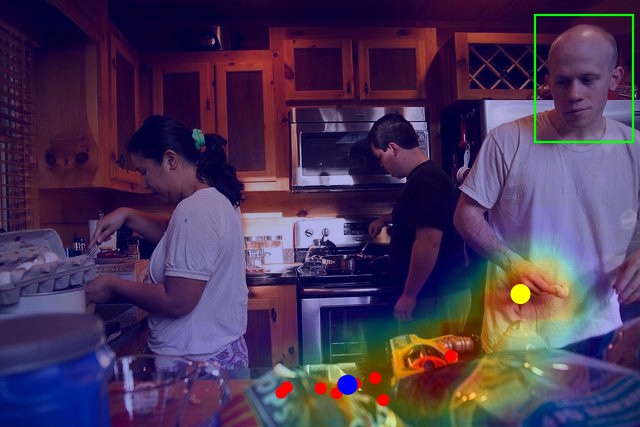} &
 \includegraphics[width=0.24\linewidth, height=0.24\linewidth]{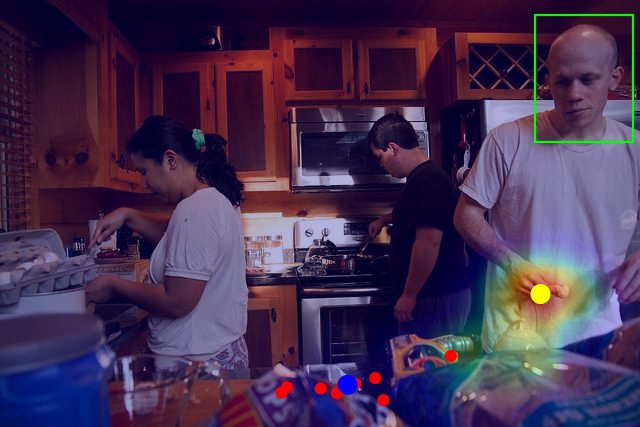}&
 \includegraphics[width=0.24\linewidth, height=0.24\linewidth]{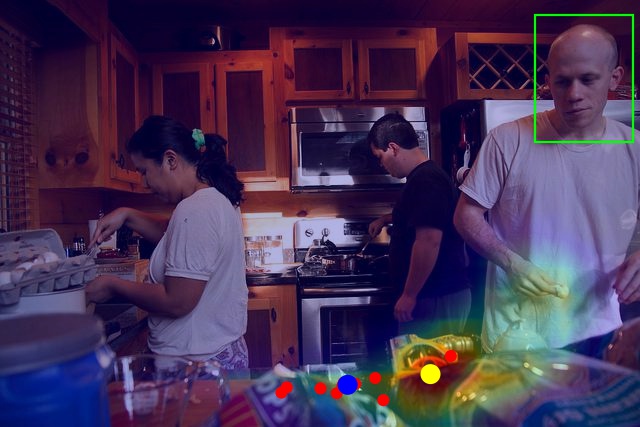}\\
  \includegraphics[width=0.24\linewidth, height=0.24\linewidth]{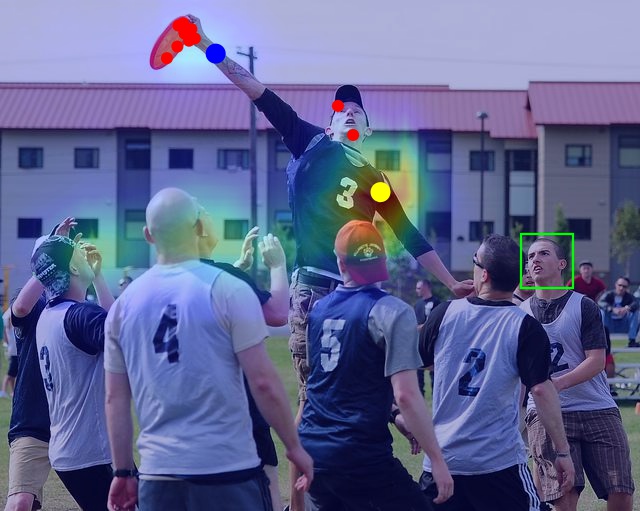} &
 \includegraphics[width=0.24\linewidth, height=0.24\linewidth]{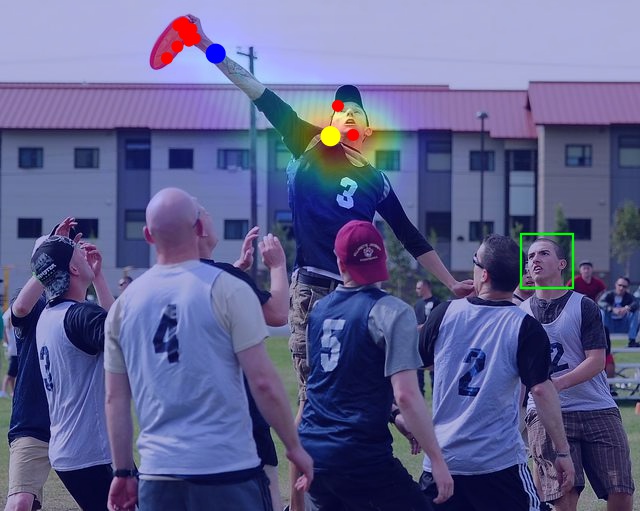} &
 \includegraphics[width=0.24\linewidth, height=0.24\linewidth]{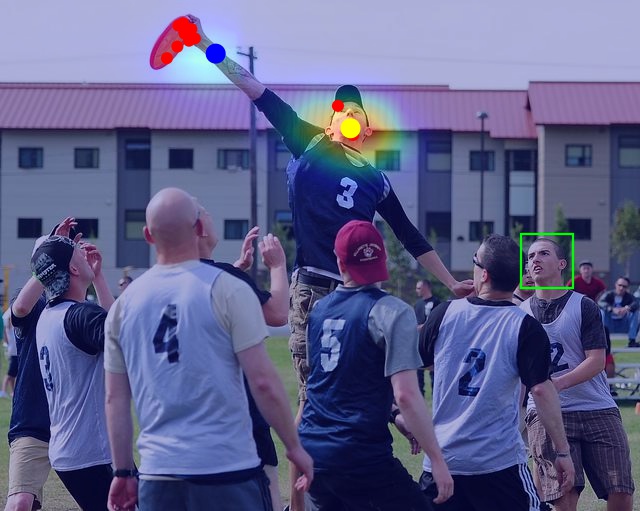}&
 \includegraphics[width=0.24\linewidth, height=0.24\linewidth]{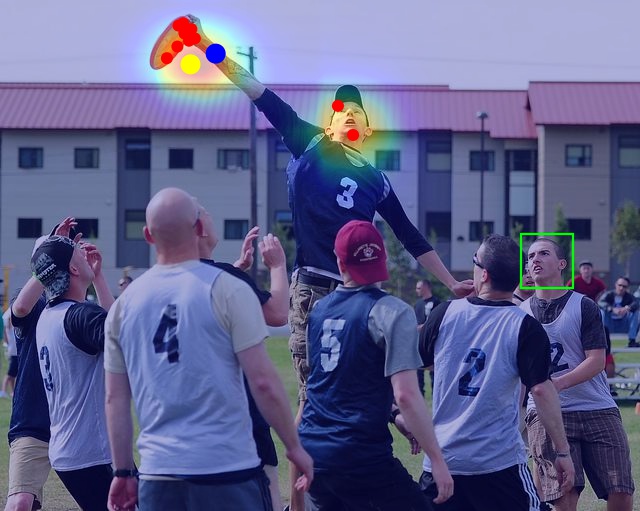}\\
  \includegraphics[width=0.24\linewidth, height=0.24\linewidth]{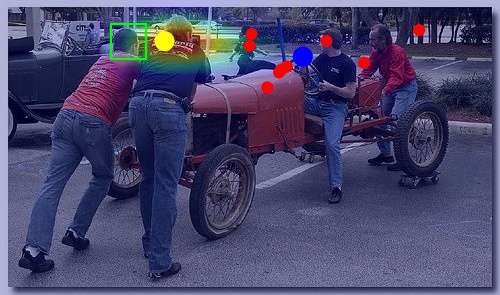} &
 \includegraphics[width=0.24\linewidth, height=0.24\linewidth]{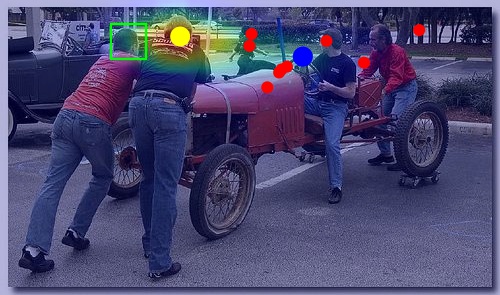} &
 \includegraphics[width=0.24\linewidth, height=0.24\linewidth]{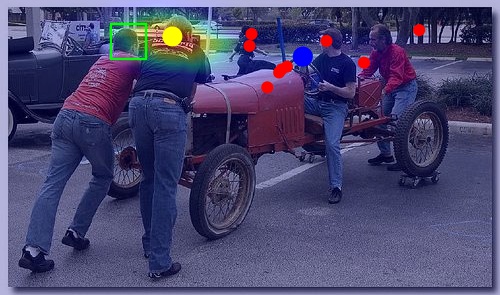}&
 \includegraphics[width=0.24\linewidth, height=0.24\linewidth]{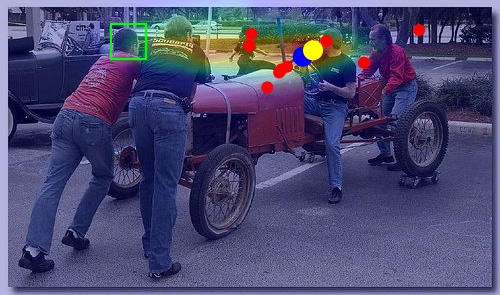}
\end{tabular}
\caption{Visualizations of the output heatmaps of our model and the VideoAtt model with and without a depth map as input. The left two columns are the model predictions without using a depth map as input, and the right two columns are the model predictions using a depth map as input. Predicted targets are plotted in yellow and the ground truth annotations are plotted in red. The average annotation is plotted in blue.}
\label{fig:4models}
\end{figure*}

In Figure \ref{fig:4models}, we visualize the outputs of our model and VideoAtt model \cite{chong2020detecting} with and without a depth map as input, to get a better understanding of the effect of the PDP method and the depth information. Our model can generate heatmap predictions better aligned with human annotations compared to the VideoAtt model, both with and without a depth map as input. This shows that the PDP task can regularize the heatmap regression task irrespective of depth information. However, without a depth map as input, it becomes more difficult for our model to predict the perfect target location. As shown in the figure, our model may predict heatmap confidence on objects inconsistent with human gaze in the depth channel (rows 1 and 4), or fail to predict on some potential gazed objects with confidence (rows 3 and 4). The depth map gives the model a much better understanding of the scene structure, making it easier for the model to infer the potential gazed objects.

\section{Example Failure Cases} \label{sec:failure_cases}
\begin{figure*}[b!]
\setlength\tabcolsep{3pt}%%
\centering
\begin{tabular}{ccc}
    \textbf{Input} &
 \textbf{Heatmap output} &
 \textbf{Target prediction}\\
 \includegraphics[width=0.26\linewidth,height=0.26\linewidth]{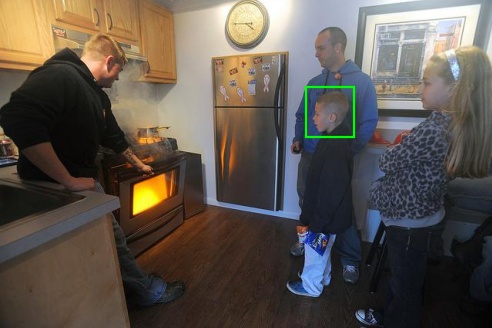} &
 \includegraphics[width=0.26\linewidth,height=0.26\linewidth]{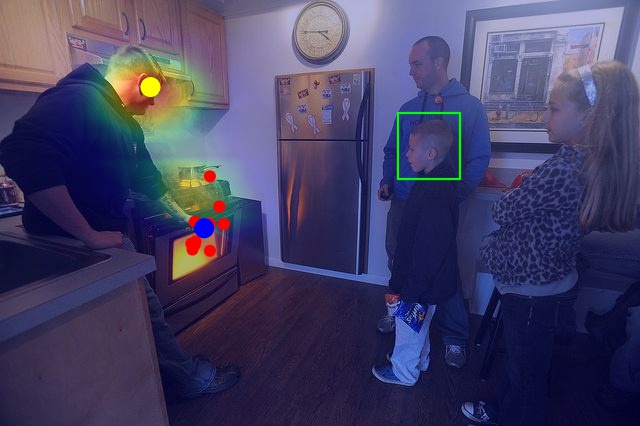} &
 \includegraphics[width=0.26\linewidth,height=0.26\linewidth]{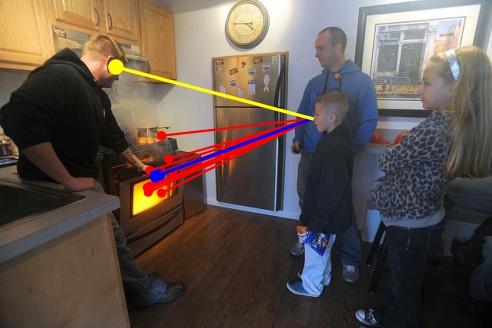}\\
 \includegraphics[width=0.26\linewidth,height=0.26\linewidth]{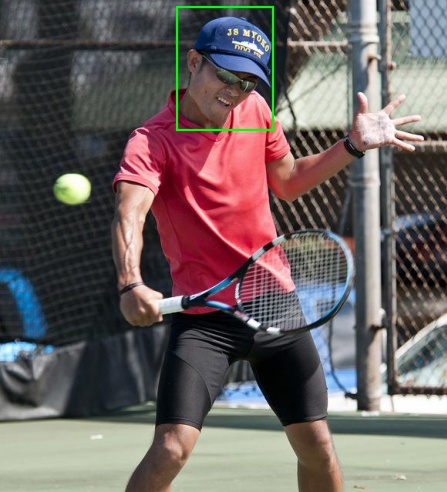} &
  \includegraphics[width=0.26\linewidth,height=0.26\linewidth]{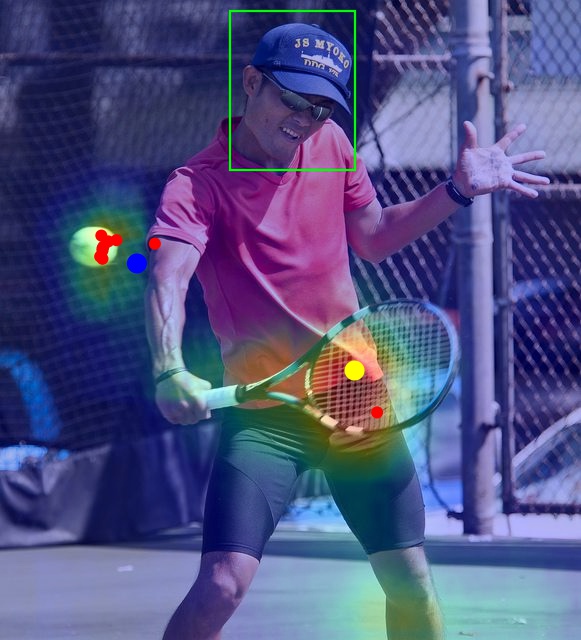} &
 \includegraphics[width=0.26\linewidth,height=0.26\linewidth]{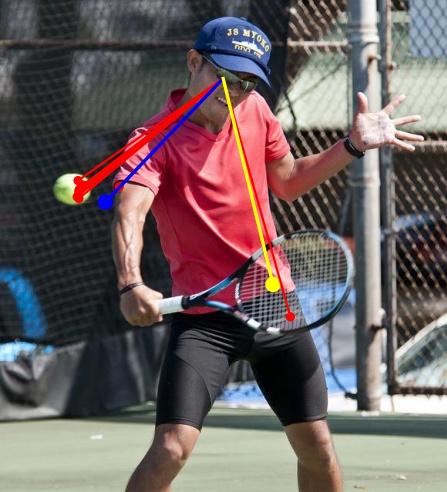}\\
 \includegraphics[width=0.26\linewidth,height=0.26\linewidth]{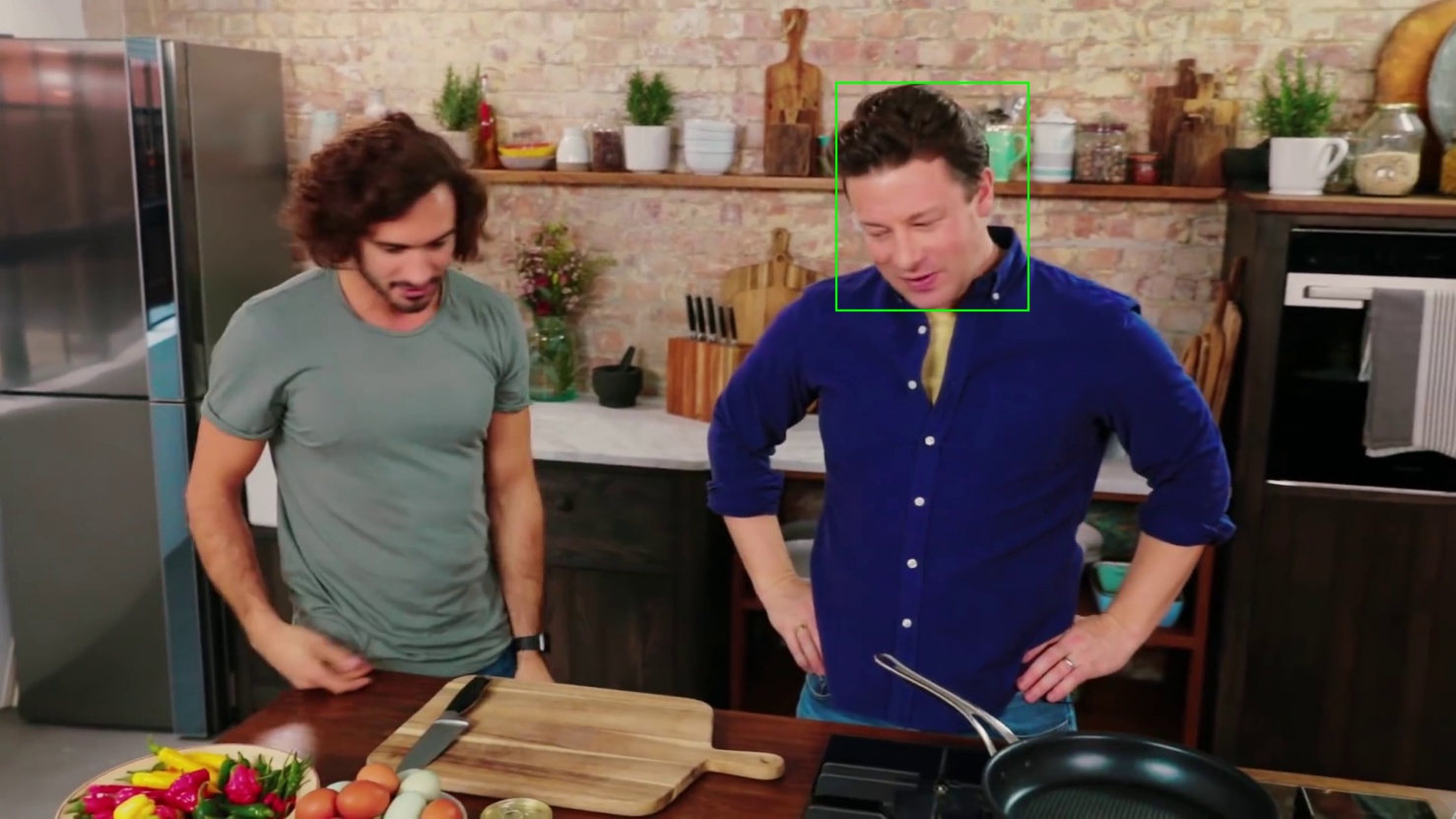} &
 \includegraphics[width=0.26\linewidth,height=0.26\linewidth]{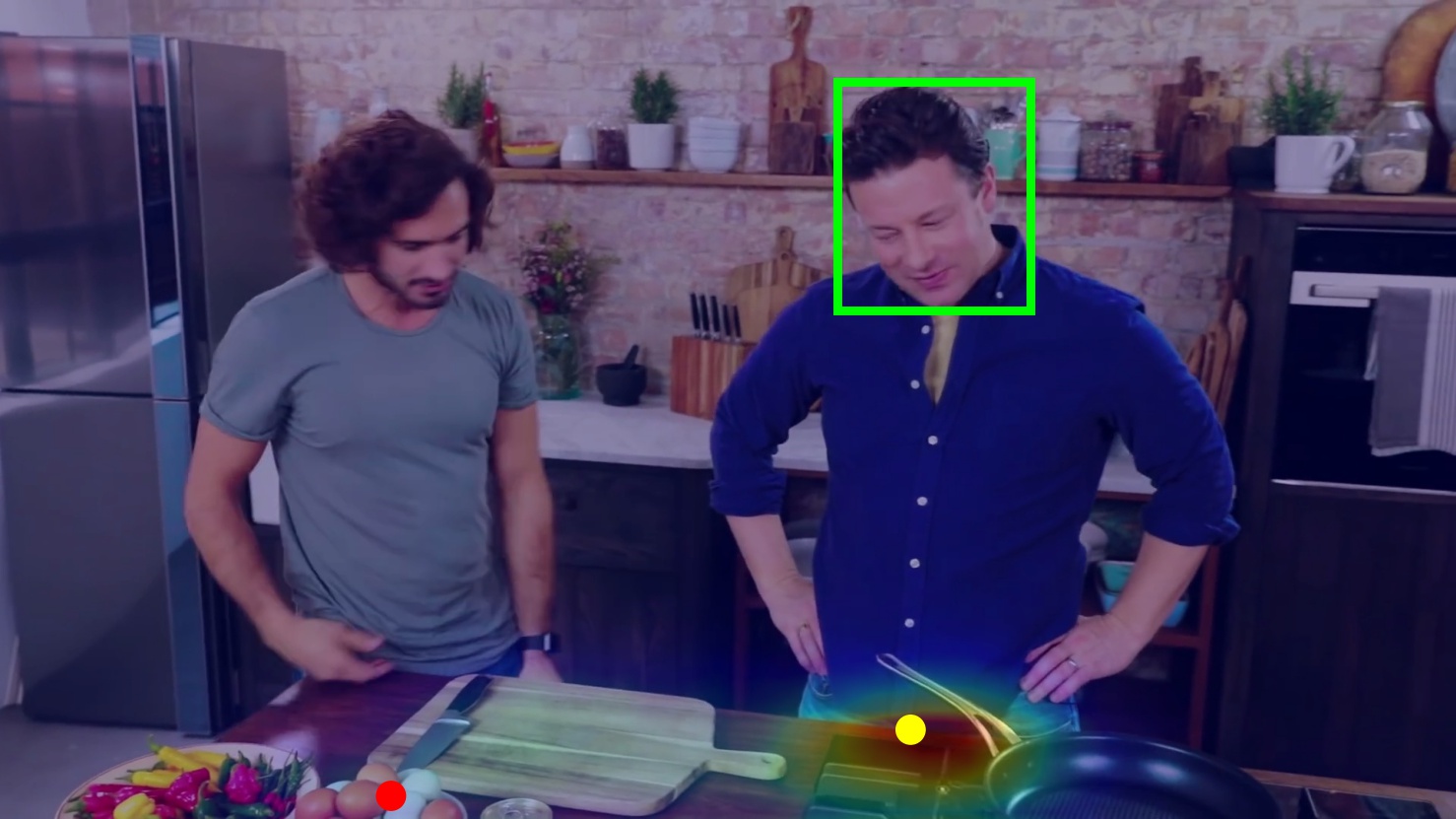} &
 \includegraphics[width=0.26\linewidth,height=0.26\linewidth]{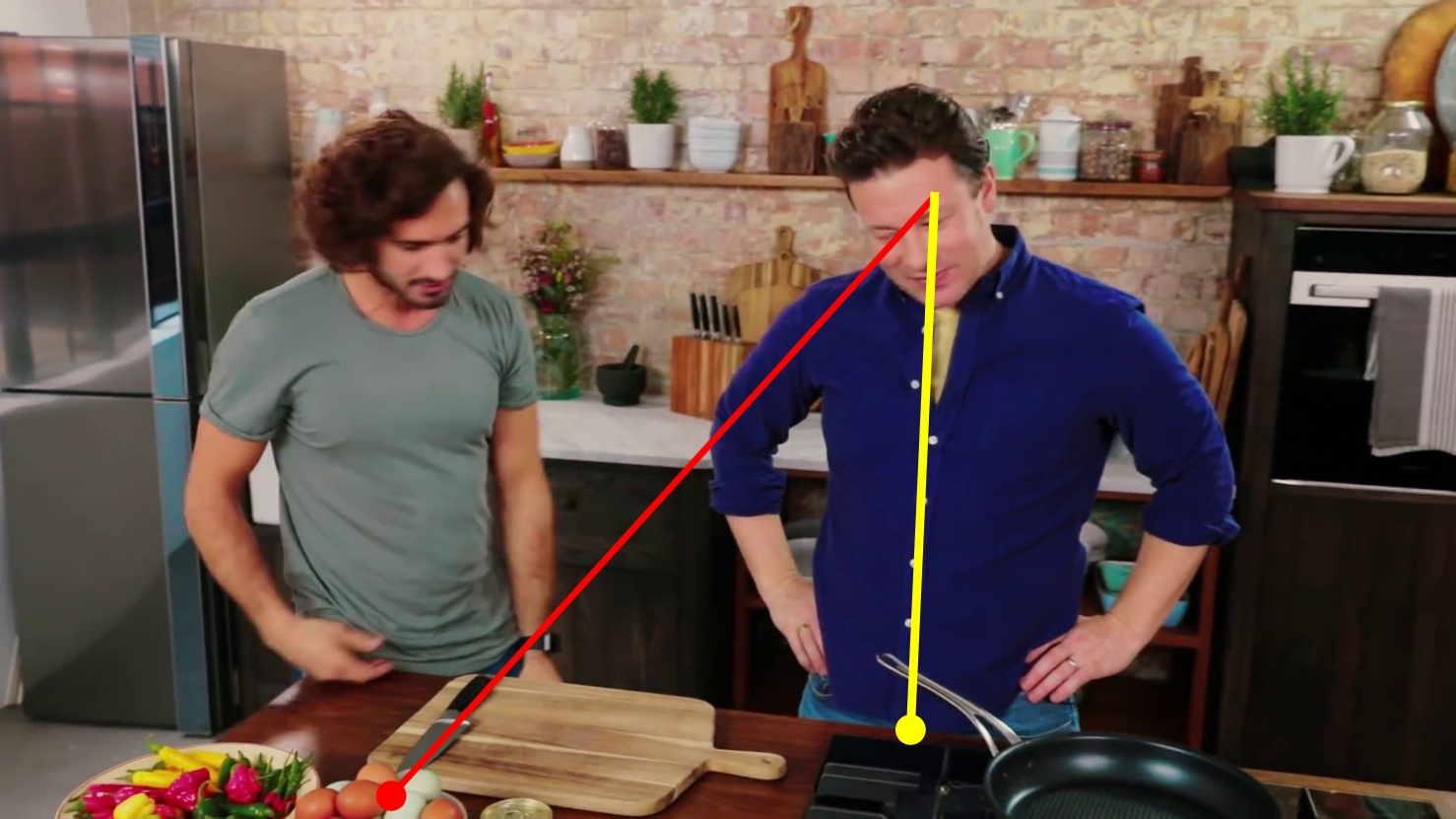}\\
  \includegraphics[width=0.26\linewidth,height=0.26\linewidth]{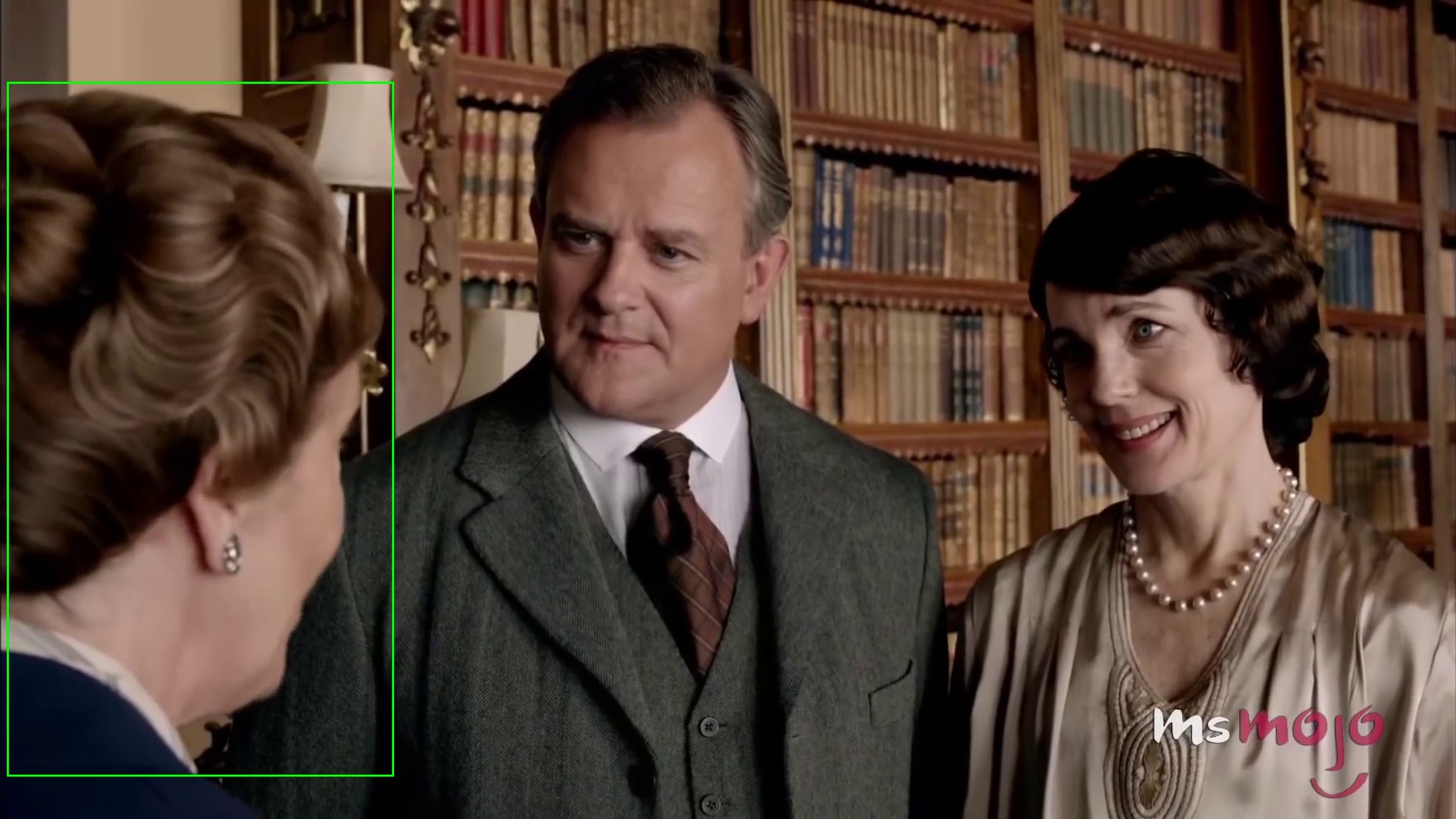} &
 \includegraphics[width=0.26\linewidth,height=0.26\linewidth]{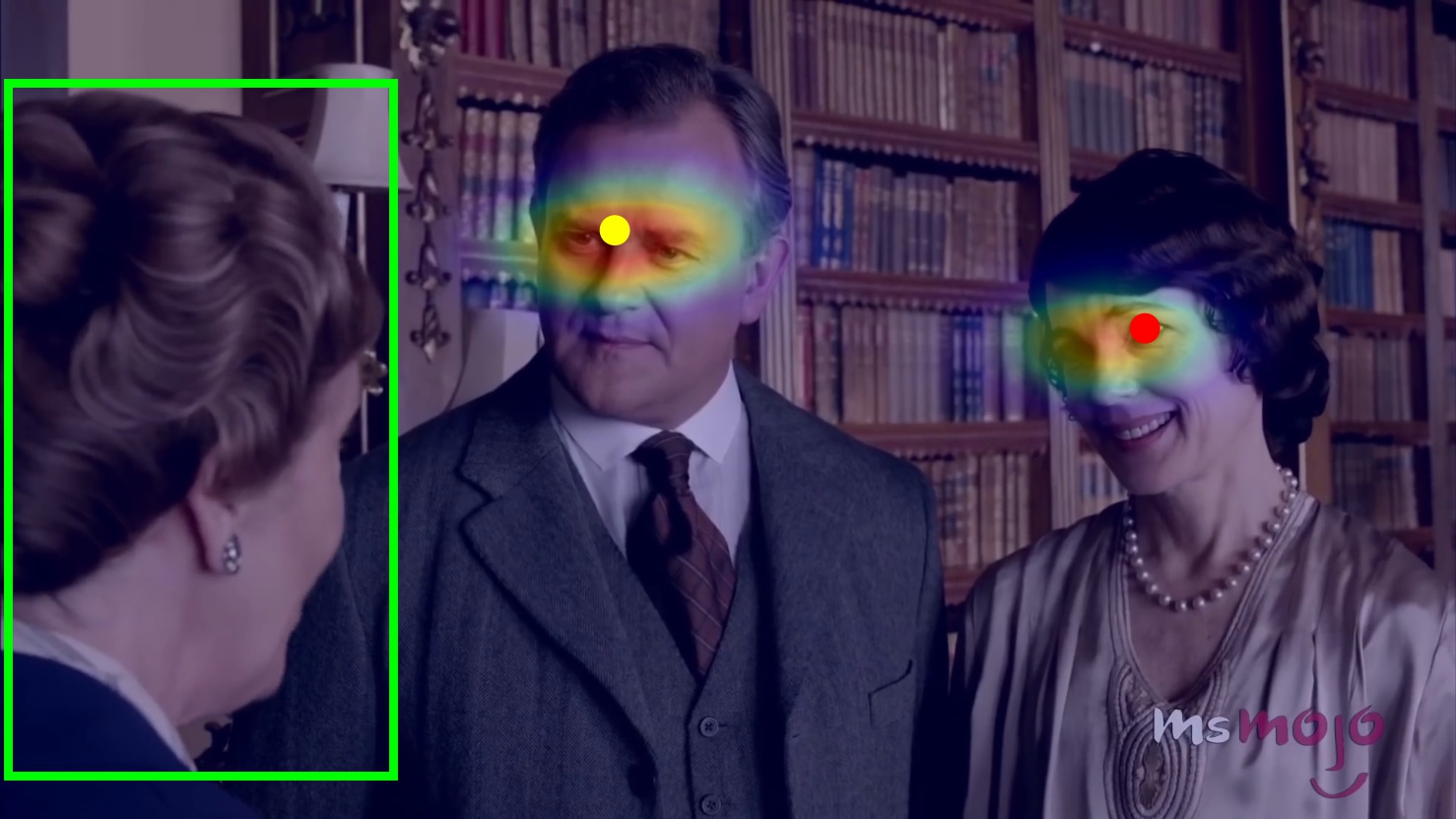} &
 \includegraphics[width=0.26\linewidth,height=0.26\linewidth]{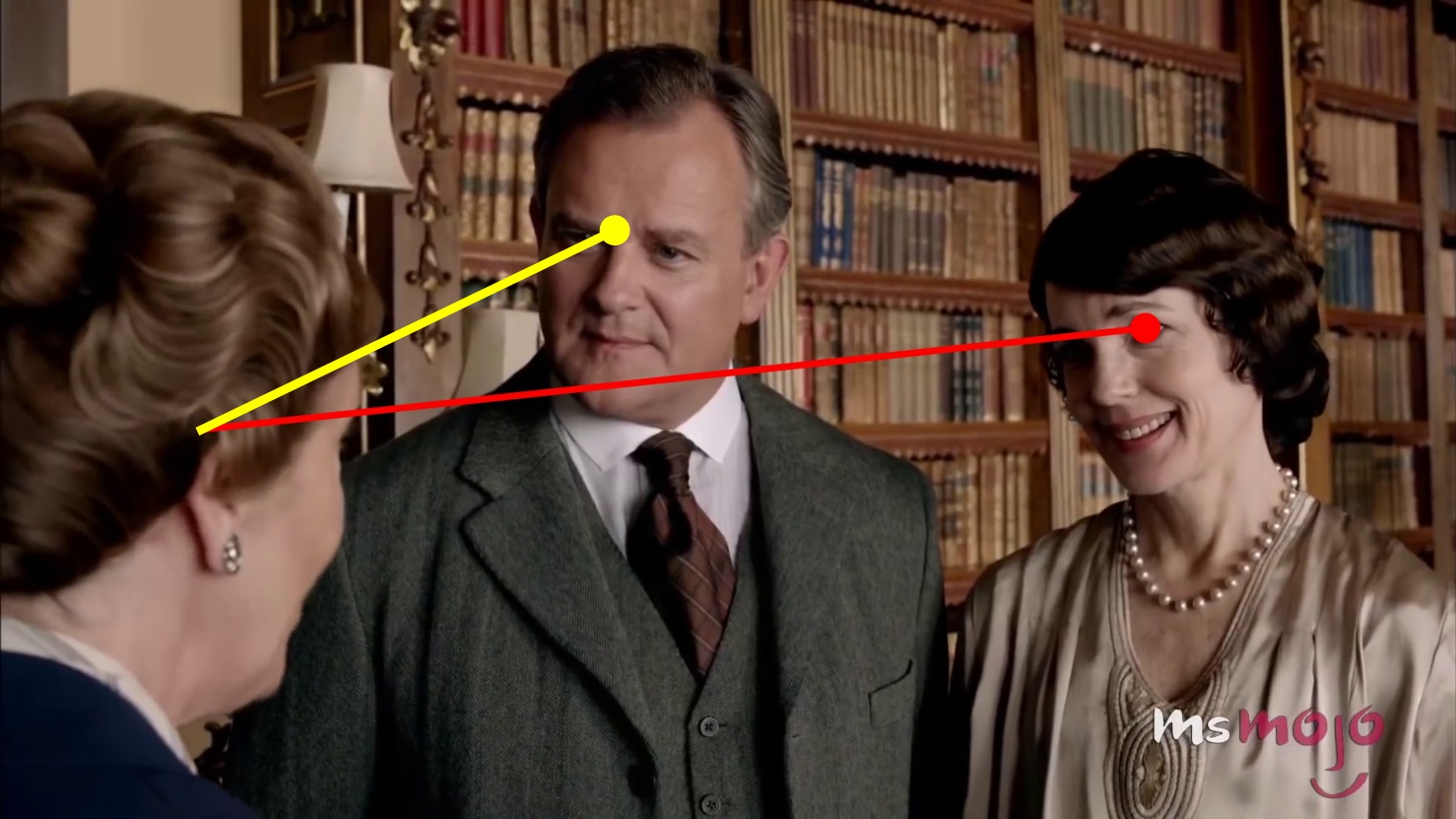}\\
\end{tabular}
\caption{Example failure cases of our model on the GazeFollow (row 1 \& 2) and VideoAttentionTarget (row 3 \& 4) datasets. Predicted targets are plotted in yellow and the ground truth annotations are plotted in red. The average annotations for images in the GazeFollow dataset are plotted in blue}
\label{fig:failure_cases}
\end{figure*}

Figure \ref{fig:failure_cases} shows some example failure cases of our model on the GazeFollow \cite{NIPS2015_ec895663} and VideoAttentionTarget \cite{chong2020detecting} datasets. Our model sometimes predicts the gaze target incorrectly when the person has a subtle eye orientation that is inconsistent with the head pose, or predicts  more confidently  a person's head instead of the actual target (row 1 and row 3). This phenomenon may be attributed to the dataset statistics that the gaze target is located on a person's head in a large number of cases. In addition, our model only takes the cropped head without extracting the cropped eye region as input, which makes it easier to employ the model but sacrifices accuracy to some extent. We would like to make our model easier to be applied to most in-the-wild data, without using a complex pre-processing step to crop the eye images, as in the DualAtt model\cite{fang2021dual}.

In some other circumstances, our model predicts multiple clusters due to the uncertainty of the input, but the predicted target determined from the maximum point in the heatmap is different from the annotation, or where most of the annotations lie, as shown in the 2nd and 4th row in Figure \ref{fig:failure_cases}. Still, our model can predict heatmap response at some level in the annotated regions. Properly speaking, our model's predictions are not totally ``wrong" in these cases and the target point determined from our heatmap still makes some sense. It is possible that the  performance of the model  increases if more annotations are obtained.

\end{document}